\documentclass[9.5pt,journal,final,finalsubmission,twocolumn]{IEEEtran}
\pdfoutput=1

\usepackage{times}
\usepackage{epsfig}
\usepackage{graphicx}
\usepackage{amsmath,amssymb} 
\usepackage{mathtools}
\usepackage{multirow}
\usepackage{subfig}
\usepackage{color}
\usepackage{rotating}
\usepackage{array}
\usepackage{cite}
\usepackage{url}
\usepackage{booktabs}
\usepackage[ruled]{algorithm}
\usepackage{algorithmic}
\usepackage{math1}

\newcommand{\bv}[1]{\mathbf{#1}}
\newcounter{proposition}

\newcommand{\ie}{i.e. }

\begin{document}
%
\title{Joint Alignment of Multiple Point Sets with Batch and Incremental Expectation-Maximization
}
%
%

\author{Georgios~D.~Evangelidis\IEEEcompsocitemizethanks{\IEEEcompsocthanksitem Georgios~D.~Evangelidis, DAQRI International, Dublin, Ireland, E-mail: georgios.evangelidis@daqri.com (This work was done while the author was with INRIA, Grenoble Rh\^one-Alpes, France)} and Radu Horaud\IEEEcompsocitemizethanks{\IEEEcompsocthanksitem Radu Horaud, INRIA Grenoble Rh\^one-Alpes, Montbonnot Saint-Martin, France, 
E-mail: radu.horaud@inria.fr}
\thanks{Funding from Agence Nationale de la Recherche (ANR) MIXCAM project \#ANR-13-BS02-0010-01 and from the European Union FP7 ERC Advanced Grant  VHIA \#340113 is greatly acknowledged.}
}

\maketitle

\begin{abstract}
This paper addresses the problem of registering multiple point sets. Solutions to this problem are often approximated by repeatedly solving for pairwise registration, which results in an uneven treatment of the sets forming a pair: a model set and a data set. The main drawback of this strategy is that the model set may contain noise and outliers, which negatively affects the estimation of the registration parameters. In contrast, the proposed formulation treats all the point sets on an equal footing. Indeed, all the points are drawn from a central Gaussian mixture, hence the registration is cast into a clustering problem. We formally derive batch and incremental EM algorithms that robustly estimate both the GMM parameters and the rotations and translations that optimally align the sets. Moreover, the mixture's means play the role of the registered set of points while the variances provide rich information about the contribution of each component to the alignment. We thoroughly test the proposed algorithms on simulated data and on challenging real data collected with range sensors. We compare them with several state-of-the-art algorithms, and we show their potential for surface reconstruction from depth data.
\end{abstract}

\begin{keywords}
Point registration, expectation maximization, mixture models, joint alignment
\end{keywords}

\section{Introduction}\label{sec:intro}

The registration of point sets is an essential methodology in computer vision, computer graphics, robotics, and medical image analysis. The vast majority of existing techniques solve the pairwise (two sets) registration problem, e.g., \cite{BeslMcKay92,Fitzgibbon2003,TsinKanade2004,MyronenkoPAMI2010,Horaud2011,JianVemuri2011}, while the multiple-set registration problem has comparatively received less attention, e.g., \cite{Williams2001,Wang2008, Govindu2014}. Solutions to this problem are often approximated by repeatedly solving for pairwise registration, either sequentially \cite{Blais1995,Masuda1995CVIU,KinectFusion1}, or via a \textit{one-versus-all} strategy \cite{Bergevin1996,castellaniCVIU2002,Huber2003}.


\begin{figure}[t]
\includegraphics[width=.99\columnwidth]{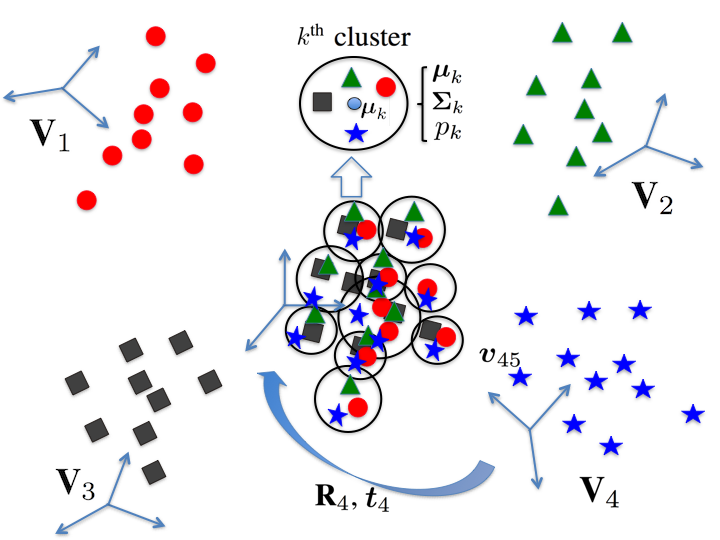}
\caption{
The proposed \textit{joint registration} method assumes that all points  from all sets, e.g. $\Vmat_1$ to $\Vmat_4$ are realizations
of the same mixture (shown in the center).
An observed point, e.g. $\vvect_{45}\in \Vmat_4$, once rotated and translated from the set-centered coordinate frame to the mixture-centered coordinate frame ($\Rmat_4$ and $\tvect_4$)
is assigned to the $k^{\text{th}}$ mixture component defined by $\muvect_k$, $\Sigmamat_k$ and $p_k$.  
As shown on the figure, the estimated
mixture is not associated to any of the point sets, as is the case with \textit{pairwise registration} methods.
}
\label{fig:gmm}
\end{figure}

Independently of the particular two-set registration algorithm that is used, the above mentioned approximate solutions have their own limitations. On the one hand, sequential strategies suffer from the well known drift accumulation owing to the chain-based optimization, i.e., sequential registration between pairs of point sets. On the other hand, one-versus-all strategies lead to a biased estimator since the registration is governed by a single reference set.  In addition, both strategies lack closed-loop information and one needs to further consider this constraint. Therefore, an unbiased solution that treats all the point sets on an equal footing and that implicitly enforces a loop constraint is particularly desirable. 

Such an unbiased solution is targeted by motion averaging approaches that build on pairwise registration schemes and aim to evenly distribute the total error across the network of point sets, either as a post-processing step~\cite{shih2008} or as an over-successive registration between pairs of point sets \cite{Govindu2014}. We rather aim to \emph{jointly} register all the point sets and not re-distribute the error from a pairwise registration. To this end, we propose a generative approach to the joint registration of multiple point sets. An arbitrary number of point sets, observed from different sensor locations, are assumed to be generated from a \textit{single} Gaussian mixture model (GMM). The problem is cast into a data clustering problem which, in turn, is solved via maximum likelihood and leads to an \emph{expectation maximization} (EM) algorithm, whereby both the mixture and registration parameters are optimally estimated. We present batch and incremental EM algorithms: both can deal with point sets of different cardinalities and contaminated by noise and outliers. 


Pairwise probabilistic registration methods constrain the GMM means to coincide with the points of one set, e.g.,~\cite{MyronenkoPAMI2010,Horaud2011}. Note that such a coincidence is inherently problematic, as long as both point sets are noisy and may include outliers. Even if one includes a uniform component in the mixture to deal with outliers~\cite{BanfieldRaftery93}, one of the sets is supposed to be ``perfect".
Instead, the means of the proposed formulation are not tight to a particular set: they result from fitting a mixture model to the data sets that are appropriately rotated and translated. In addition to registration, this also achieves scene reconstruction, since the cluster means may be viewed as the scene model. The proposed formulation implicitly enforces a closed-loop constraint. In other words, the proposed model assumes a star network topology, while the pairwise registration schemes assume a ring topology or a fully connected network 


This article is an extended version of  \cite{Evangelidis-ECCV-2014}. Several aspects of the proposed model are discussed into more detail, namely initialization, behavior, complexity and advantages over existing methods. In addition to the batch EM described in \cite{Evangelidis-ECCV-2014}, we introduce an incremental version of EM, which solves the parameter estimation problem more efficiently at the cost of less accurate results. Experiments with novel datasets and benchmarks with several recent methods are included as well.

The remainder of this paper is organized as follows: Section~\ref{sec:related_work} discusses the related work. Section~\ref{sec:problem_formulation} formulates the problem in a generative probabilistic framework. Section~\ref{sec:jrmpc} describes the batch EM together with an algorithm analysis while the incremental version is presented in Section~\ref{sec:online-jrmpc}. Section~\ref{sec:initialization} describes in detail various initialization procedures. Section~\ref{sec:experiments} presents the experimental results and Section~\ref{sec:conclusions} concludes the paper.\footnote{Matlab code, datasets and videos are available at \url{https://team.inria.fr/perception/research/jrmpc/}.}


\section{Related work}\label{sec:related_work}

The two-set registration problem is usually solved by ICP~\cite{BeslMcKay92,Chen1992} or by one of its variants~\cite{chetverikov2002trimmed,Fitzgibbon2003,TsinKanade2004,Segal2009, goicp2013}. While ICP alternates between hard assignments and transformation estimation, more sophisticated registration approaches replace the binary assignments with probabilities~\cite{Wells97,GrangerPennec2002,CR:CVIU03,MyronenkoPAMI2010,Horaud2011}. Nevertheless, whether based on ICP or on soft assignments,  these methods consider one set as the ``model'' and the other set as the ``data'', thus leading to solutions that are biased as long as both sets may contain noise and outliers. Alternatively, \cite{JianVemuri2011,Hermans2011} consider two Gaussian mixtures, one per point set, and the rigid transformation is applied to one of these mixtures. This leads to a non-linear optimization problem. 

Multiple point set registration is often addressed by a sequential pairwise registration strategy~\cite{Blais1995,Masuda1995CVIU,Chen1992}, in particular when an online solution is required. Whenever an additional set is available, the model set is updated using either an ICP-like or a probabilistic scheme. Apart from the drawbacks associated with pairwise registration, this incremental mode of operation is subject to error propagation, while it fails to close any existing loop.  As for offline applications, several approaches have been proposed, being mostly based on the underlying network, a.k.a. viewgraph, defined by the sets (represented as nodes) and their relative overlap (represented as edges). The majority of these methods initialize the poses via a pairwise registration.

The first solution for the problem in question was proposed in~\cite{Bergevin1996}, where the sets are organized in a star-shaped network with one of the sets in the center, and such that any two sets are linked via two edges, hence by combining two rigid transformations. An algorithm computes the transformations incrementally based on a point-to-plane ICP algorithm~\cite{Chen1992}. \cite{Benjemaa1998} proposed to accelerate this algorithm by allowing incremental updates once pairwise registration within the loop has been performed. \cite{Huber2003} starts with pairwise registrations to build the set graph, while a global registration step eliminates inconsistent matches and leads to the model graph whereby poses are provided. All these methods, however, consider in practice one set as reference, thus favoring a biased and non-symmetric solution. 

Alternatively, \cite{masuda2002registration} proposes a method to register multiple range images based on shape modeling: a point-to-surface distance is defined, the signed distance field, and the algorithm alternates between alignment and registration. Measurement errors and wrong correspondences are handled by a robust loss function. This method is well suited for dense range data since a surface representation is necessary.

Other methods consider known and fixed correspondences across multiple sets, thus updating only the transformations to balance the global error over the viewgraph~\cite{Williams2001,castellaniCVIU2002,Sharp2004,Krishnan2007,shih2008,Torsello2011}. The main principle of these methods is that transformations along a network cycle ideally compose to the identity transformation. The cycles may refer to either minor loops between two adjacent sets or a larger cycle over the network.\footnote{When a spanning tree is used, an unused edge is added to obtain a cycle.} Provided an approximate alignment, the goal is to minimize the on-cycle accumulated error from registering pairs of relevant (nearby) views. However, when data are ignored, a low inconsistency between coordinate frames does not necessarily mean better surface registration, in particular when good initialization is not available. As a consequence, these methods just ``spread'' any existing bias across the network without any correspondence refinement. 

An alternative approach consists of considering a dense sequence of depth images and of estimating slight misalignments between these images. If the images are linearly correlated, the image alignment can be obtained via low-rank decomposition of a large matrix which has as columns the misaligned images. This formulation has been successfully applied to 2D image alignment \cite{peng2012rasl} and extended to align images gathered with a RGB-D sensor \cite{thomas2012robust}. The method is however limited to small camera motions such as to preserve the necessary condition that the images are linearly correlated. Our method addresses a different scenario, because it can handle large camera displacements and it does not necessitate dense RGB-D data. We conclude that our method and  \cite{thomas2012robust} are complementary.

Several recent methods built on the motion averaging principle introduced in~\cite{Govindu2001} and based on rotation averaging \cite{Hartley2013}.
Provided the view network, \cite{Govindu2014} suggests a motion averaged ICP algorithm. This algorithm alternates between the correspondence step and a double motion update. Any edge of the network implies an ICP run that updates a relative motion, and the redundancy information from \emph{all} the relative motions in turn lead to a new global motion (one transformation per set) through the Lie-algebraic motion averaging principle. Then, the global motion information is back propagated in order to re-update the relative motions in a globally consistent manner. Again, the main assumption behind averaging is that traversing a cycle on the view network implies no motion. However, point correspondences are also updated here. \cite{ZhongyuLi2014} adopts the same technique but it employs trimmed-ICP~\cite{ChetverikovStepanovKrsek2005} to compute pairwise motions. Note that an existing closed-loop may need to be pre-defined or pre-detected.

Probabilistic methods have been also proposed. As in~\cite{JianVemuri2011}, \cite{Wang2008} represents each point set as a GMM and the non-rigid transformations are applied to cluster centers rather than to raw points. The model parameters are estimated by minimizing the Jensen-Shannon divergence of multiple densities and a probabilistic mean shape is built (as a by-product) from the convex combination of the aligned sets. This method vitally depends on each set's clusters, thus requiring highly and well structured point sets with no outliers. More closely to our method, \cite{Goldberger1999}  proposed an EM algorithm that alternates between the reconstruction of the object's mean shape and the registration between the sets and this shape. Despite the same principle, i.e., an emerging mean shape generates the sample sets, \cite{Goldberger1999} considers \emph{given correspondences} as well as several simplifications. KinectFusion~\cite{KinectFusion1} would roughly fall into this category owing to its model-to-frame registration strategy. Unlike these approaches, \cite{Cui2013} and \cite{Mateo2014} build on pairwise registrations. The former generalizes \cite{MyronenkoPAMI2010} to align multiple super-resolved depth images by jointly optimizing many pairwise alignments along with compensating for pixel-dependent systematic bias. The latter extends the objective function of~\cite{Williams2001} and \cite{Krishnan2007} by considering correspondences as missing data that are inferred along with the pairwise transformations in an EM fashion. Recently, \cite{DanelljanCVPR2016} proposed an extension of~\cite{Evangelidis-ECCV-2014} that integrates RGB information which enables better initial matches. In a large-scale outdoor context, \cite{Shiratori2015} exploits positioning and map data to a pre-detect closed loop, while it proposes a multiple point set extension of \cite{Segal2009} to simultaneously refine the intra-loop poses of the range sensor.

\section{Problem formulation}\label{sec:problem_formulation}

Let $\Vmat_j=[\vvect_{j1} \ldots \vvect_{ji}  \ldots \vvect_{jN_j}] \in \mathbb{R}^{3\times N_j}$ be $N_j$ data points that belong to point set  $j$ and let $M$ be the number of point sets. We denote with $\Vmat =\{\Vmat_j\}_{j=1}^M$ the union of all these sets. 
A rigid transformation $\phi_j:\mathbb{R}^3\rightarrow\mathbb{R}^3$, \ie a rotation matrix and a translation vector, maps $\vvect_{ji}$ from a \textit{set-centered} frame to a \textit{model-centered} frame, such that all the points form all the sets are expressed in the same coordinate frame. 
The objective is to estimate the $M$ data-set-to-model-set transformations under the assumption that the observed points are generated from the same mixture model
\begin{equation}\label{eq:mixture_equation}
  P(\vvect_{ji})
 = \sum_{k=1}^{K} p_k\mathcal{N}\big(\mathbf{R}_j \vvect_{ji} + \tvect_j ; \boldsymbol{\mu}_k,\mathbf{\Sigma}_k\big) + p_{K+1} \mathcal{U}(h),
\end{equation}
where  $\mathbf{R}_j\in\mathbb{R}^{3\times 3}$ is a rotation matrix and $\tvect_j\in\mathbb{R}^3$ is a translation vector such that $\phi_j (\vvect_{ji}) = \mathbf{R}_j \vvect_{ji} + \tvect_j$, $p_k$ are the mixing coefficients with $\sum_{k=1}^{K+1} p_k=1$, $\boldsymbol{\mu}_k\in\mathbb{R}^3$ and $\mathbf{\Sigma}_k\in\mathbb{R}^{3\times 3}$ are the mean vectors and covariance matrices respectively, and $\mathcal{U}(h)$ is the uniform distribution parameterized by the volume $h$ of the 3D convex hull encompassing the data \cite{Horaud2011}.
We now define $\gamma$ as the ratio between outliers and inliers
\begin{equation}\label{eq:new_constraint}
\gamma = \frac{p_{K+1}}{\sum\limits_{k=1}^K p_k}.
\end{equation}
This allows to balance the outlier/inlier proportion via a judicious choice of $\gamma$. To summarize, the model parameters are
\begin{equation}\label{eq:parameter_set}
\Theta=\left\{ \{p_k,\boldsymbol{\mu}_k,\mathbf{\Sigma}_k\}_{k=1}^K, \{ \mathbf{R}_j,\tvect_j \}_{j=1}^M\right\}.
\end{equation}
This problem can be solved using an EM algorithm. We define hidden variables $\mathcal{Z} = \{Z_{ji}|j \in [1\hdots M]$, $i\in [1\hdots N_j]\}$ such that $Z_{ji}=k$ means that observation $\vvect_{ji}$ is assigned to the $k$-th component of the mixture, and we seek to estimate the parameters $\Theta$ by
maximizing  the expected complete-data log-likelihood given the observed data
\begin{align} 
\label{eq:ECDLL}
\mathcal{E}(\Theta | \Vmat,\mathcal{Z} )&= 
\mathbb{E}_{\mathcal{Z}}[\log \mathit{P}(\Vmat,\mathcal{Z}| \Vmat;\Theta)] \nonumber \\
&=\sum_{\mathcal{Z}}P(\mathcal{Z}|\Vmat;\Theta)\log(P(\Vmat, \mathcal{Z};\Theta)).
\end{align}

\section{Batch Registration}\label{sec:jrmpc}

Assuming that the observed data are independent and identically distributed, it is straightforward to write (\ref{eq:ECDLL}) as
\begin{equation} \label{eq:complete_likelihood_new}
\mathcal{E}(\Theta | \Vmat,\mathcal{Z} ) = 
\sum\limits_{j,i,k} 
 \alpha_{jik} \Big(\log p_k + \log P(\vvect_{ji}|Z_{ji}=k;\Theta)\Big),
\end{equation}  
where $\alpha_{jik}=P(Z_{ji}=k|\vvect_{ji};\Theta)$ are the posteriors. 
By replacing the standard expressions of the likelihoods~\cite{Bishop2006} and by ignoring constant terms, (\ref{eq:complete_likelihood_new}) can be written as an objective function of the form
\begin{align} \label{eq:objective_function}
f(\Theta)= & -\frac{1}{2} 
\sum\limits_{j,i,k}
\alpha_{jik} \big(\|\phi_j(\vvect_{ji}) - \boldsymbol{\mu}_k)\|^2_{\bv{\Sigma}_k}
+\log|\bv{\Sigma}_k|  \nonumber \\
&-2\log p_k\big)+\log p_{K+1}\sum\limits_{j,i} \alpha_{ji(K+1)},
\end{align}
where $|\cdot|$ denotes the determinant and $\|\yvect\|^2_{\mathbf{A}}=\yvect^\top \mathbf{A}^{-1}\yvect$. 
The model is farther restricted to isotropic covariances, \ie $\bv{\Sigma}_k=\sigma_k\bv{I}_3$, since this leads to closed-form maximization solutions for all the model parameters~\eqref{eq:parameter_set}, while non-isotropic covariances lead to a more complex convex optimization problem with no significant gain in accuracy~\cite{Horaud2011}.
Particular care must be given to the estimation of the rotation matrices, namely a constrained optimization problem
\begin{equation}
\label{eq:constrained_optimization}
\begin{cases}
\max_{\Theta}  f(\Theta) \\
\text{s.t. }  \quad \bv{R}_j^\top\bv{R}_j=\bv{I}_3 \text{ and } |\bv{R}_j|=1, \forall j\in [1\hdots M],
\end{cases}
\end{equation}
which can be solved via EM.
Notice that the standard M steps for Gaussian mixtures are augmented with a step that estimates the rigid transformation parameters. We will refer to this algorithm as \textit{joint registration of multiple point clouds} (JRMPC). The batch version will be referred to as JRMPC-B and it is outlined in Algorithm~\ref{alg:JR-MPC}. 
This leads to a conditional maximization procedure~\cite{MengRubin93}. 
Each M-step first estimates the transformation parameters, given the current responsibilities and Gaussian mixture parameters, and then estimates the new mixture parameters, given the new transformation parameters. It is of course possible to adopt a reverse order, in particular when rough rigid transformations are available. However, the proposed order does not assume such prior information.


\subsection{E-step}
The posterior probability of point $\vvect_{ji}$ to be associated with cluster $k$, e.g. an inlier, is 
\begin{equation}\label{eq:posterior_final}
\alpha_{jik}=\frac{\beta_{jik} } 
{\sum\limits_{s=1}^{K} \big( \beta_{jis}
\big) + \frac{\gamma}{h(\gamma +1)}},
\end{equation}
where $\gamma/h(\gamma+1)$ accounts for the uniform component in the mixture, and with the notation:
\begin{equation}\label{eq:posterior_notation}
\beta_{jik} = \frac{p_k}
{\sigma_k^{3/2}} \exp{\bigg( \frac{- \| \bv{R}_j \vvect_{ji} + \tvect_j - \boldsymbol{\mu}_k\|^2}{2\sigma_k}} \bigg).
\end{equation}
Therefore, the posterior probability of being an \emph{outlier} is simply given by 
$\alpha_{ji\:K+1} = 1-\sum_{k=1}^K\alpha_{jik}$.
As shown in Algorithm~\ref{alg:JR-MPC}, the posterior probability at the $q$-th iteration, $\alpha_{jik}^q$, is computed from (\ref{eq:posterior_final}) using the parameter set $\Theta^{q-1}$.\\

\begin{algorithm}[t!]
\algsetup{linenosize=\scriptsize}
\caption{Batch Joint Registration of Multiple Point Clouds (JRMPC-B)}
\begin{algorithmic}[1]
\REQUIRE Initial parameter set $\Theta^0$, number of components $K$, number of iterations $Q$.
\STATE $q\leftarrow 1$
\REPEAT
\item[]  \hspace{-0.2cm}\emph{E-step}:
\STATE Use $\Theta^{q-1}$ to estimate posterior probabilities
$\alpha_{jik}^q = P(Z_{ji}=k|\vvect_{ji};\Theta^{q-1})$, \ie \eqref{eq:posterior_final}.
\item[]  \hspace{-0.2cm}\emph{M-rigid-step}:
\STATE  Use $\alpha_{jik}^q$, $\boldsymbol{\mu}_k^{q-1}$ and $\mathbf{\Sigma}_k^{q-1}$ to estimate $\mathbf{R}_j^{q}$ and $\tvect_j^{q}$, \ie \eqref{eq:optimum_rotation} and \eqref{eq:optimum_translation}.
\item[]  \hspace{-0.2cm}\emph{M-GMM-step}:
\STATE  Use $\alpha_{jik}^q$, $\mathbf{R}_j^{q}$ and $\tvect_j^{q}$ to estimate the means $\boldsymbol{\mu}_k^{q}$, \ie \eqref{eq:gmm_mean}.
\STATE  Use $\alpha_{jik}^q$, $\mathbf{R}_j^{q}$, $\tvect_j^{q}$ and $\boldsymbol{\mu}_k^{q}$ to estimate the covariances $\mathbf{\Sigma}_k^{q}$, \ie  \eqref{eq:gmm_variance}.
\STATE  Use $\alpha_{jik}^q$ to estimate the priors $p_k^{q}$, \ie \eqref{eq:lagrange_mutliplier}.
\STATE $q\leftarrow q+1$
\UNTIL $q>Q$ (or $\Theta$'s update is negligible) 
\RETURN $\Theta^{q}$
\end{algorithmic}\label{alg:JR-MPC}
\end{algorithm}

\subsection{M-rigid-step}
This step estimates the rotations $\bv{R}_j$ and translations $\tvect_j$ that maximize $f(\Theta)$, given current values for $\alpha_{jik}$, $\boldsymbol{\mu}_k$, $\bv{\Sigma}_k$, and $p_k$. Notice that this estimation can be carried out independently for each set $\Vmat_j$. 
By setting the GMM parameters to their current values, we reformulate the problem of estimating the rotations and translations. 
The rigid transformation parameters that maximize $f(\Theta)$ can be estimated from the following constrained minimization 
\begin{equation}\label{eq:rotation_optimization}
\begin{cases}
\displaystyle\min_{\bv{R}_j,\tvect_j} \|(\bv{R}_j\Wmat_j +\tvect_j\evect^\top - \Mmat)\bv{\Lambda}_j\|^2_F & \\ 
\text{s.t.} \quad \bv{R}_j^\top\bv{R}_j=\bv{I}_3 \mbox{ and } |\bv{R}_j|=1,&
\end{cases}
\end{equation}
where $\bv{\Lambda}_j \in \mathbb{R}^{K\times K}$ is a diagonal matrix with entries $\lambda_{jkk}=(\sum_{i=1}^{N_j}\alpha_{jik}/ \sigma_k)^{1/2} $, $\Mmat = [\boldsymbol{\mu}_1, \ldots, \boldsymbol{\mu}_K] \in \mathbb{R}^{3\times K}$, $\evect \in \mathbb{R}^{K}$ is a vector of ones, $\|\cdot\|_F$ denotes the Frobenius norm, and $\Wmat_j=[\wvect_{j1},\ldots,\wvect_{jK}]  \in \mathbb{R}^{3\times K}$, where $\wvect_{jk}$ is the weighted average of the $j$-th point set assigned to the $k$-th mixture component
\begin{equation}\label{eq:virtual-points}
\wvect_{jk} = \frac{\sum_{i=1}^{N_j}\alpha_{jik}\vvect_{ji}}{\sum_{i=1}^{N_j}\alpha_{jik}}, 
\end{equation}
The minimization \eqref{eq:rotation_optimization} can be solved in closed-form and is a weighted version of the solution~\cite{Umeyama91}. 
The optimal rotation matrices are
\begin{equation}\label{eq:optimum_rotation}
\bv{R}_j = \bv{U}_j^l\bv{S}_j{\bv{U}_j^r}^\top, \; \forall j\in[1\dots M],
\end{equation}
where $\bv{U}_j^l$ and $\bv{U}_j^r$ are the left and right matrices respectively, obtained from the singular value decomposition of matrix $\Mmat \bv{\Lambda}_j\bv{P}_j\bv{\Lambda}_j\bv{W}_j^\top$, with
\begin{equation}
\bv{P}_j = \bv{I}_3-\frac{\bv{\Lambda}_j\evect \evect^\top \bv{\Lambda}_j}{\evect^\top\bv{\Lambda}_j^2\evect}
\end{equation} 
is a projection matrix and
$\bv{S}_j= \text{diag}(1,1,|\bv{U}_j^l||\bv{U}_j^r|)$.
Once the optimal rotation matrices are estimated, the optimal translation vectors are easily computed with
\begin{equation}\label{eq:optimum_translation}
\tvect_j  = \frac{1}{\text{trace}(\bv{\Lambda}_j^2)}(\Mmat-\bv{R}_j \Wmat_j)\bv{\Lambda}_j^2\evect,\; \forall j\in[1\dots M].
\end{equation}
Note that each rigid transform $\phi_j$ aligns the GMM means with K virtual points $\{\wvect_{jk}\}_{k=1}^K$ (one virtual point per component). Therefore, the proposed method can deal with point sets of different cardinalities and the number of components in the mixture, $K$, can be chosen independently of these cardinalities. This is an important advantage over pairwise registration methods that assume that the cardinalities of the two point sets must be similar.

\subsection{M-GMM-step}
Given rigid transformation estimates and posterior probabilities, one can use standard optimization techniques to compute the optimal means and covariances:
\begin{align}
\label{eq:gmm_mean}
\boldsymbol{\mu}_k  &= \frac{\sum\limits_{j=1}^{M} \sum\limits_{i=1}^{N_j} \alpha_{jik}  (\bv{R}_j \vvect_{ji}+\tvect_j )}
{\sum\limits_{j=1}^{M} \sum\limits_{i=1}^{N_j} \alpha_{jik} }, \\
\label{eq:gmm_variance}
{\sigma_k}  &= \frac{\sum\limits_{j=1}^{M} \sum\limits_{i=1}^{N_j} \alpha_{jik}
\| \bv{R}_j \vvect_{ji}+\tvect_j - \boldsymbol{\mu}_k  \|_2^2}{3 \sum\limits_{j=1}^{M} \sum\limits_{i=1}^{N_j} \alpha_{jik}} +\epsilon^2,
\end{align}
where $\epsilon$ is a small scalar to avoid singularities. As for the priors, we introduce a Lagrange multiplier to take into account the constraint $\sum_{k=1}^{K+1} p_k=1$. This leads to the following dual function
\begin{align}
g (p_1,\ldots,p_{K},\eta) = & \sum\limits_{k=1}^K \bigg(\log p_k\sum\limits_{j=1}^{M} \sum\limits_{i=1}^{N_j}\alpha_{jik} \bigg) \nonumber \\
+  & \eta \bigg( \sum\limits_{k=1}^{K} p_k -\frac{1}{1+\gamma}\bigg).
\end{align}
and its optimization yields
\begin{align}\label{eq:lagrange_mutliplier}
p_k  &= \frac{1}{\eta}\sum\limits_{j=1}^{M} \sum\limits_{i=1}^{N_j}\alpha_{jik}, \;\; \forall k \in \{1,\ldots, K\}\\
p_{K+1}  &= 1 - \sum\limits_{k=1}^Kp_k , \label{eq:langrange_multiplier2},
\end{align}
with $\eta =(\gamma+1)(N-\sum_{j=1}^{M} \sum_{i=1}^{N_j}\alpha_{ji\:K+1})$
and 
$N=\sum_{j=1}^{M} N_j$. 
Note that if $\gamma\to 0$, which means that there is no uniform component in the mixture, then $\eta \to N$, which is in agreement with~\cite{Bishop2006}.

\subsection{Algorithm Analysis}\label{sec:complexity}

The leading complexity of JRMPC-B is $O(NK)$ owing to E-step and equation \eqref{eq:posterior_notation}. If $\bar{N}$ is the average cardinality of a point set, the complexity can be written as $O(\bar{N}MK)$. Typically, $K<\bar{N}$ owing to underlying clustering while $K$ could be close to or even greater than $\bar{N}$  when many non-ovelapping sets cover a large volume.

The proposed algorithm has a number of advantages over pairwise registration methods. Such methods, e.g.~\cite{Williams2001,Govindu2014, ZhongyuLi2014}, are intrinsically more time-consuming than joint registration because one has to consider all point-set-to-point-set combinations. Either using EM or ICP when registering each pair of sets, the evaluation of all point-to-point distances is needed. Such a strategy requires $O(\bar{N}^2M^2)$ operations in principle. This complexity can be decreased by structuring the data, e.g. using KD-trees, at the cost of data structure building. Approximate solutions, e.g., sequential or one-versus-all approaches, consider $M-1$ pairs of sets, thus requiring $O(\bar{N}^2M)$ operations at the expense of performance.

Another important difference between joint and pairwise registrations is that the former puts all the point sets on an equal footing and registration is truly cast into clustering, i.e. \eqref{eq:mixture_equation}, whereas the latter performs an unbalanced treatment of the point sets, i.e. one set constitutes the data and the other set constitutes the model. More precisely, when EM is used for registering pairs~\cite{MyronenkoPAMI2010, Horaud2011}, the generative model $\mathcal{N}\big(\mathbf{R}_j \vvect_{ji} + \tvect_j ; \boldsymbol{\mu}_k,\mathbf{\Sigma}_k\big)$ in \eqref{eq:mixture_equation} is replaced with $\mathcal{N}\big( \vvect_{aj} ; \mathbf{R}_{ab} \vvect_{bi}+ \tvect_{ab} ,\mathbf{\Sigma}_{bi}\big)$, where $\vvect_{ai}$ belongs to point set $a$ (the data), $\vvect_{bj}$ belongs to point set $b$ (the model) and $(\mathbf{R}_{ab},\tvect_{ab})$ is the rigid alignment that maps $b$ onto $a$.  Hence, in the joint case, the mixture is modeled by a set of free parameters, while in the pairwise case, the means directly depend on the rigid parameters. The immediate consequence is that noisy points or outliers that may be present in the data will propagate via this dependency and will give rise to bad means in the mixture. When ICP is used, again one of the sets is identified with the model.

The minimal configuration required by the proposed method consists of two sets with at least three overlapping points. The algorithm can be applied to a large number of point sets. However in this case, the computation time increases linearly with $N$ on the premise that $K$ does not depend on the number of point sets to be aligned. For this reason, it is interesting to provide an incremental version of the algorithm, on the following ground: once $M$ point sets ($M\geq 2$) are aligned with the JRMPC-B algorithm described above, new sets can be added incrementally (one at a time) and aligned with the current model, at a lower computational cost than the batch algorithm, using update formulae for mixture parameters. The incremental version of the algorithm is described in detail in the next section.

\section{Incremental Registration}\label{sec:online-jrmpc}
The incremental version of the proposed registration method, referred to as JRMPC-I,  is outlined in Algorithm~\ref{alg:JRMPC-I}. This algorithm considers the new $m$-th point set to be aligned with $m-1$ already registered sets. The latter and the corresponding $m-1$ transformations are not updated and therefore are not used as input and output arguments. The JRMPC-I algorithm starts with computing the responsibilities $\alpha_{mik}$, it then estimates the rigid transformation that aligns the set with the already aligned sets, $\bv{R}_m$ and $\tvect_m$, and finally updates the mixture parameters. This process can be optionally repeated for a few iterations ($Q$). While the mixture parameters are initialized with those previously calculated, the integration of the new set requires initialization of $\bv{R}_m$ and $\tvect_m$, referred to as $\bv{R}_m^0$ and $\tvect_m^0$ in Algorithm~\ref{alg:JRMPC-I}. One possible strategy that has been successfully used with a moving RGB-D camera, e.g., \cite{KinectFusion1}, is to initialize the rigid transformation with $\bv{R}_m^0 = \bv{R}_{m-1}$ and $\tvect_m^0 = \tvect_{m-1}$. In the more general case, one can use the initialization strategy discussed in Section~\ref{sec:initialization}.

\begin{algorithm}[t!]
\algsetup{linenosize=\scriptsize}
\caption{Incremental Joint Registration of Multiple Point Clouds (JRMPC-I)}
\begin{algorithmic}[1]
\REQUIRE GMM parameters estimated with JRMPC-B, $\{p_k^{1:m-1}, \boldsymbol{\mu}_k^{1:m-1}, \sigma_k^{1:m-1}\}_{k=1}^K$,  $m$-th point set, rigid transformation $\bv{R}_m^0,\tvect_m^0$, number of iterations $Q$.
\STATE $q\leftarrow 1$
\REPEAT
\item[]  \hspace{-0.2cm}\emph{E-step}:
\STATE Use  \eqref{eq:posterior_m} to estimate $\alpha_{mik}^q$, $1\leq i \leq N_m, 1\leq k \leq K$.
\item[]  \hspace{-0.2cm}\emph{M-rigid-step}:
\STATE Use  \eqref{eq:optimum_rotation} and \eqref{eq:optimum_translation} to estimate $\mathbf{R}_m^{q},\tvect_m^{q}$.
\item[]  \hspace{-0.2cm}\emph{M-GMM-step}:
\STATE Use  \eqref{eq:update_mean_integration} to update ${\boldsymbol{\mu}_k^{1:m}}^q, 1\leq k \leq K$.
\STATE Use \eqref{eq:update_variance_integration} to update ${\sigma_k^{1:m}}^q, 1\leq k \leq K$.
\STATE Use  \eqref{eq:update_prior_integration}  to update ${p_k^{1:m}}^q, 1\leq k \leq K$.
\STATE $q\leftarrow q+1$
\UNTIL $q>Q$.
\RETURN $\mathbf{R}_m^{q}, \tvect_m^{q}, \{{p_k^{1:m}}^q, {\boldsymbol{\mu}_k^{1:m}}^q, {\sigma_k^{1:m}}^q\}_{k=1}^K$
\end{algorithmic}\label{alg:JRMPC-I}
\end{algorithm}


We denote with $\{p_k^{1:m-1}, \boldsymbol{\mu}_k^{1:m-1}, \sigma_k^{1:m-1}\}_{k=1}^K$ the GMM parameters estimated with JRMPC-B, where the notation ${1:m}$ denotes the sets from $1$ to $m$. The incremental registration algorithm proceeds iteratively. The E-step computes the responsibilities associated with the $m$-th set:
\begin{equation}
\label{eq:posterior_m}
\alpha_{mik}=\frac{\beta_{mik} } 
{\sum\limits_{s=1}^{K} \big( \beta_{mis}
\big) + \frac{\gamma}{h(\gamma +1)}},
\end{equation}
\begin{equation*}
\beta_{mik} = \frac{p_k^{1:m-1}}
{({\sigma_k^{1:m-1}})^{\frac{3}{2}}}\exp{\bigg( \frac{-\| \bv{R}_m \vvect_{mi} -\tvect_m  - \boldsymbol{\mu}_k^{1:m-1}\|^2}{2\sigma_k^{1:m-1}}} \bigg).
\end{equation*}

The M-rigid-step uses equations \eqref{eq:virtual-points}-\eqref{eq:optimum_translation} to calculate $\bv{R}_m$ and $\tvect_m$ in closed-form. This rigid transformation aligns the $m$-th set with the GMM means that explain the previously aligned sets, hence the joint alignment of all the sets. 
The M-GMM-step updates the means, covariances and priors:
\begin{align}
\label{eq:update_mean_integration}
\boldsymbol{\mu}_k^{1:m} &= \frac{ \zeta_{k} \boldsymbol{\mu}_k^{1:m-1 }+\uvect_{mk} }{\zeta_{k}+1},\\
\label{eq:update_variance_integration}
\sigma_k^{1:m} &= \frac{ \zeta_{k}  \sigma_k^{1:m-1}  + \|\Delta\boldsymbol{\mu}_k\|^2 
 -\Delta\boldsymbol{\mu}_k^{\top}(\uvect_{mk} -\frac{1}{\alpha_{mk}}\boldsymbol{\mu}_k^{1:m-1})}
 {\zeta_{k} +1}, \\
\label{eq:update_prior_integration}
p_k^{1:m}  &=  \frac{\alpha_{mk}\zeta_{k} + 1}{\eta^{1:m}},
\end{align}
with
\begin{align*}
  \uvect_{mk} &= \bv{R}_m  \wvect_{mk} + \tvect_m ,\quad \Delta\boldsymbol{\mu}_k = \boldsymbol{\mu}_k^{1:m} -\boldsymbol{\mu}_k^{1:m-1},\\
\zeta_{k} &= \frac{\eta^{1:m-1} p_k^{1:m-1} }{ \alpha_{mk}},\quad \alpha_{mk} = \sum_{i=1}^{N_m} \alpha_{mik}, \\
\eta^{1:m} &= \eta^{1:m-1} + (\gamma+1)(N_m+1-\alpha_{mk}).
\end{align*}

The number of iterations that JRMPC-I needs to converge depends on its initialization. It was noticed that a small number of iterations are sufficient when data are gathered from a smoothly moving camera. 
Once a few sets have been integrated with JRMPC-I, it may be useful to run JRMPC-B in order to obtain a globally optimal alignment and to reject the outliers. Also, it is worthwhile to remark that JRMPC-I is not meant to grow the model, i.e. it is not designed to increase the number of components of the Gaussian mixture as point sets, possibly with no overlap, are incrementally added. 
JRMPC-I should be merely used when an efficient algorithm is needed. In the particular case of a large number of sets, e.g., depth sequences, a temporally hierarchical scheme that benefits from both versions is recommended to cope with the large memory requirements.

\section{Initialization}\label{sec:initialization}
It is well known that initialization plays a crucial role in EM procedures. Therefore, we discuss here initialization options well suited for point set registration.
We assume no prior information about the position and orientation of the camera(s) with respect to the scene. However, information such as the calibration parameters of a network of static cameras, or transformations between pairs of point sets, could be used if available. We also assume that there is sufficient overlap between pairs of point sets. The sensitivity of our method to the amount of overlap is tested an analyzed in Sec.~\ref{sec:experiments}.

When the point sets have a sufficient joint overlap, 
the translation vectors can be initialized by centroid differences, i.e., $\tvect_j^0=\bar{\boldsymbol{\mu}}-\bar{\vvect}_j$, where $\bar{\boldsymbol{\mu}}$ is the centroid of the cluster centers and $\bar{\vvect}_j$ the centroid of the $j-$th set. If the point sets suffer from strong artifacts, e.g., flying pixels, the difference of medians can be used instead. Rotation matrices can be simply initialized with $\mathbf{R}_j^0=\mathbf{I}_3$. 
Instead, when many non-overlapping pairs exist, a pairwise registration is preferred, i.e., the minimum number of pairs that leads to a rough global alignment can be  registered beforehand.

Several strategies may be adopted for initializing the mixture parameters. One way to do it is to initialize the means with the points of one set.  Another way is to distribute the means on the surface of a sphere that encompasses the convex hull of the point sets already centered at $\bar{\boldsymbol{\mu}}$. Concerning the variance, we found that starting with a high value yields very good results and that the variances quickly converge to the final values. Our algorithm converges much faster than EM algorithms that adopt a deterministic annealing behavior, i.e. the variance is decreased according to an annealing schedule.  
Finally the priors are initialized with $1/(K+1)$, where we remind that $K$ is the number of Gaussian components. While update formulae for the priors are provided with both our algorithms, in practice it was found that keeping the priors constant affect neither the convergence nor the quality of the registrations. 
Notice that any rough pre-alignment of the sets results in a very good initilazation of the mixture parameters, that is,  the means can be intitialized from re-sampling the registered set while the variances can be intiialized such that each cluster encompasses a sufficient number of points. 

In order to choose the number of components, we propose the following empirical strategy. If the cardinalities of the point sets are similar, one can use $K=\bar{N}$ (recall that $\bar{N}$ is the average number of points in a set). However, $K$ may be chosen to be smaller than $\bar{N}$ if the sets highly overlap, or larger if there are many non-overlapping sets. One should notice that the number of components in the mixture merely depends on the data and on the application at hand. Experimentally we found that $K\ll\bar{N}$ yields excellent alignment results in the presence of dense depth data, as provided by depth sensors.

\begin{figure*}
\begin{tabular}{cccc}
\includegraphics[width=.31\textwidth]{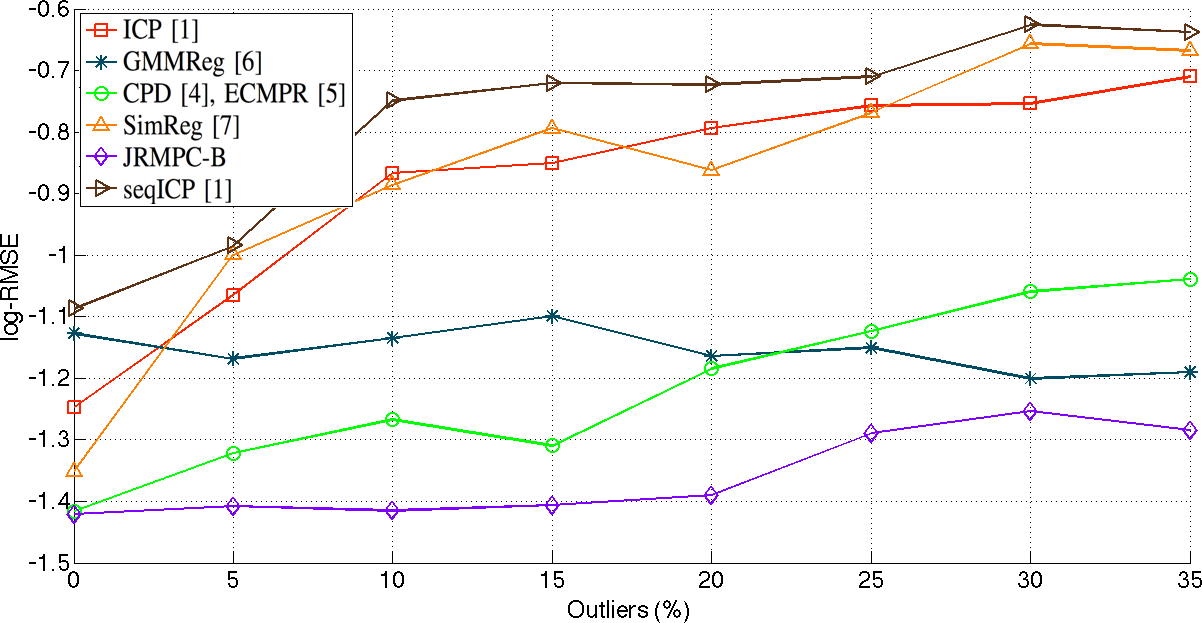}
& \includegraphics[width=.31\textwidth]{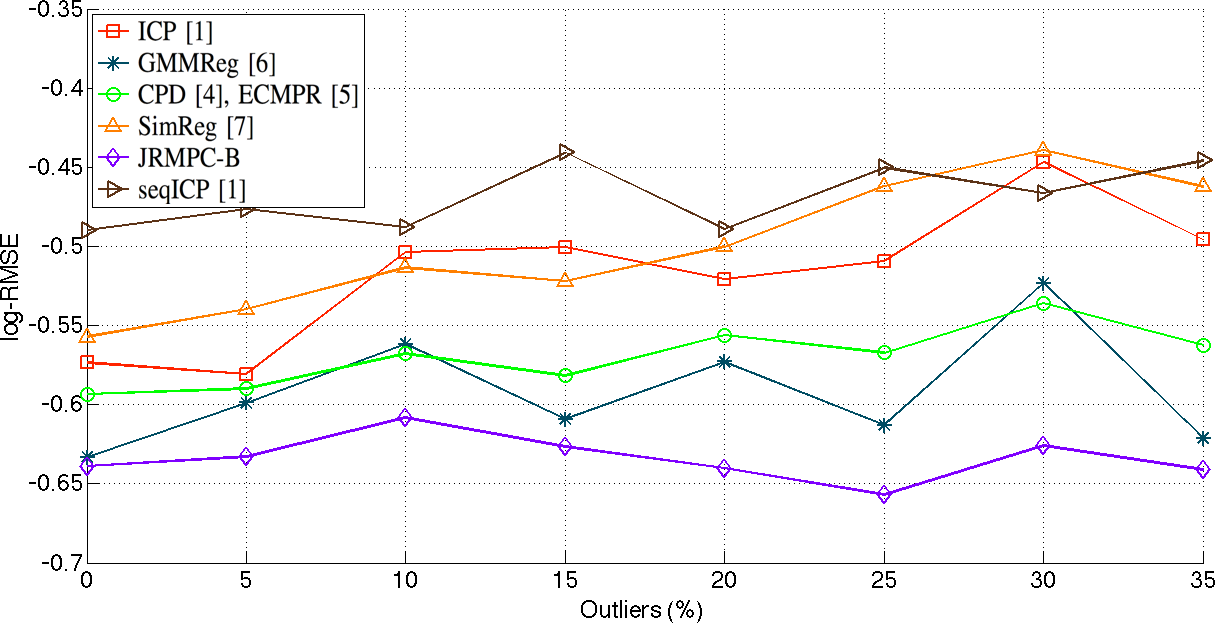}
& \includegraphics[width=.31\textwidth]{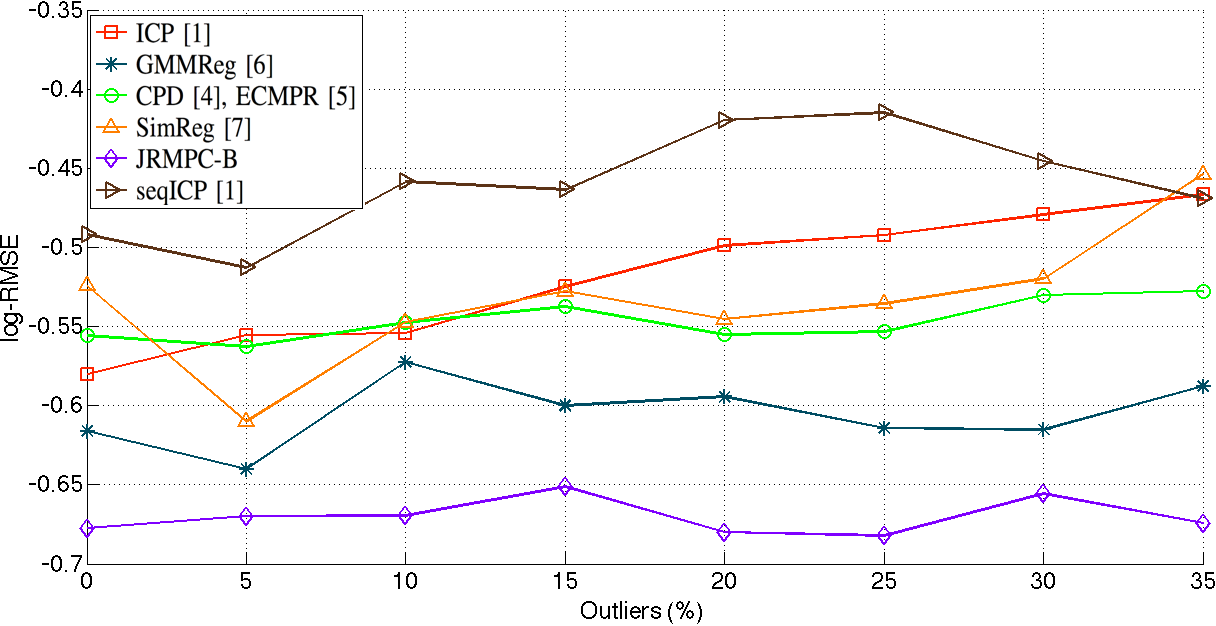}\\
\includegraphics[width=.31\textwidth]{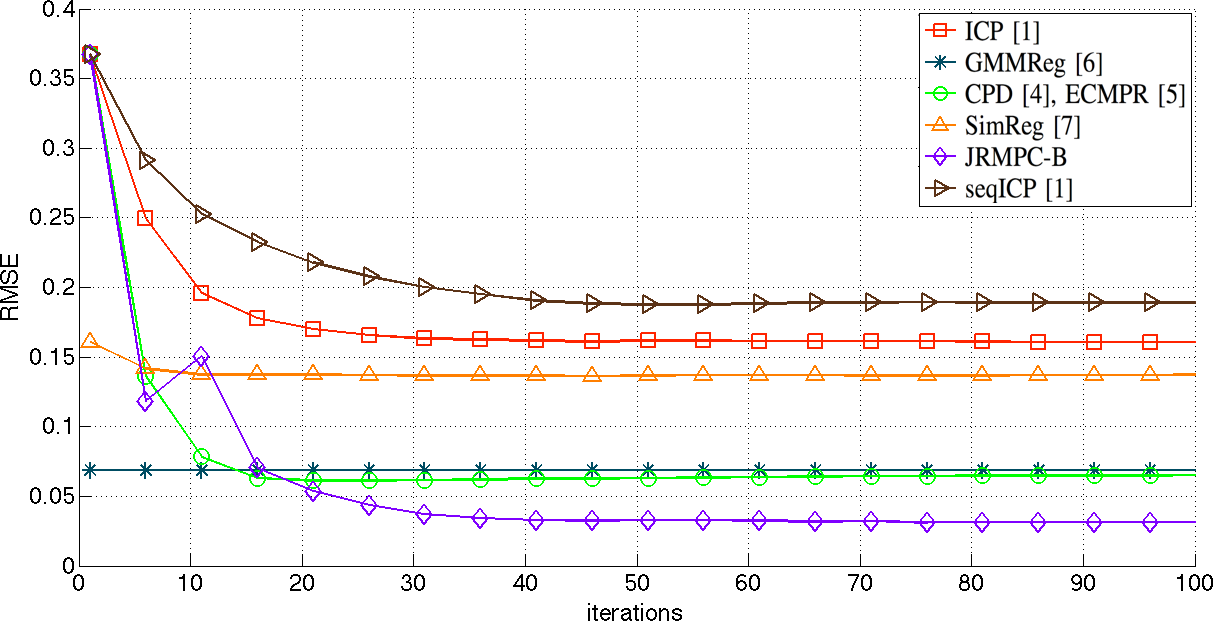}
&\includegraphics[width=.31\textwidth]{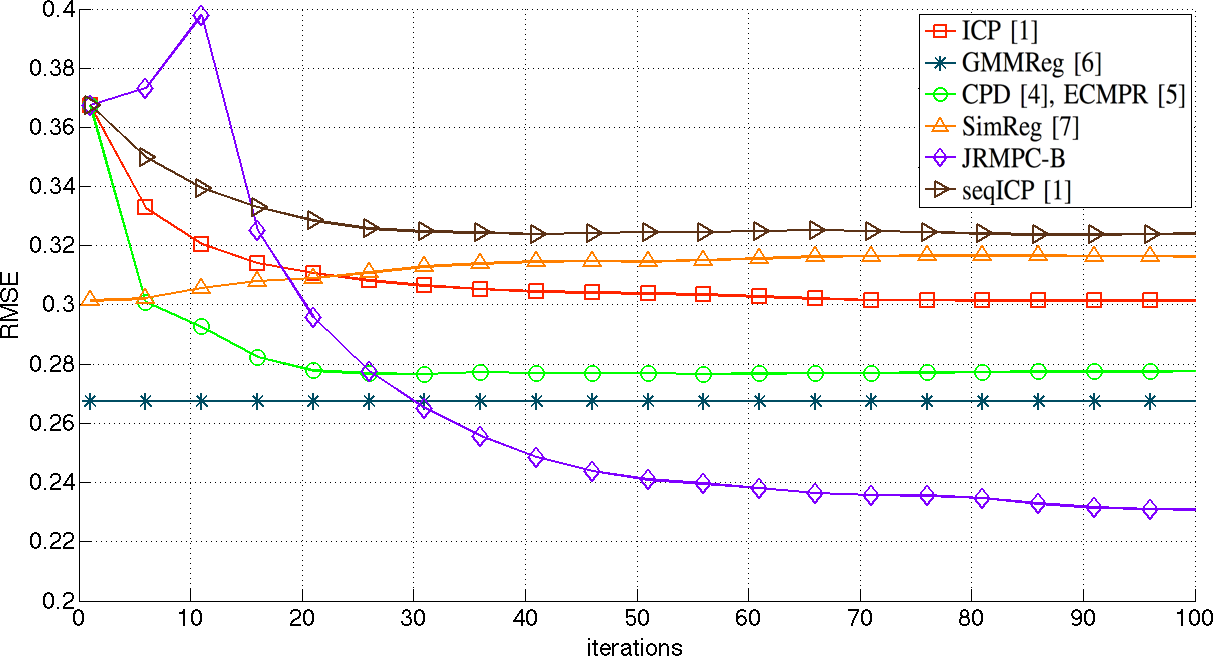}
&\includegraphics[width=.31\textwidth]{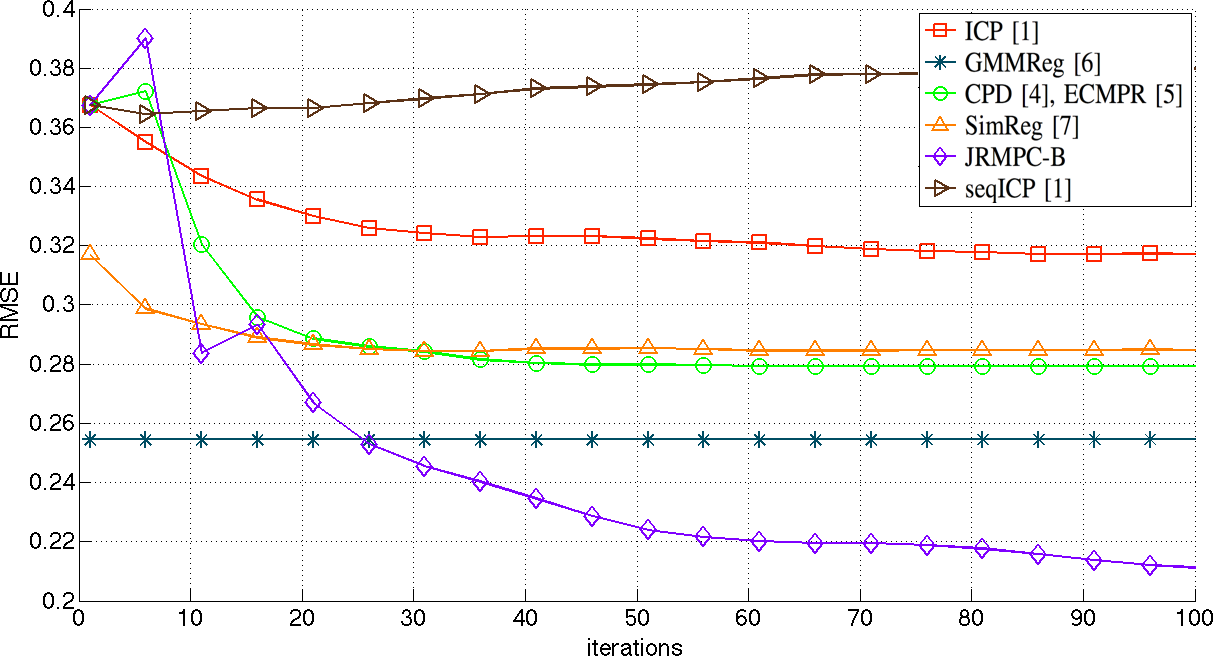}\\
(a) & (b) & (c)
\end{tabular}
\caption{\emph{Top:} $\log$-RMSE as a function of outlier percentage when SNR=$10dB$.  \emph{Bottom:} The learning curve of algorithms for a range of $100$ iterations when the models are disturbed by SNR=$10dB$ and $20\%$ outliers.  (a) ``Lucy'', (b) ``Bunny'' (c) ``Armadillo''.}
\label{fig:dragon_noise_and_outliers}
\end{figure*}

\begin{figure*}
\centering
\begin{tabular}{cccc}
\includegraphics[width=.23\textwidth]{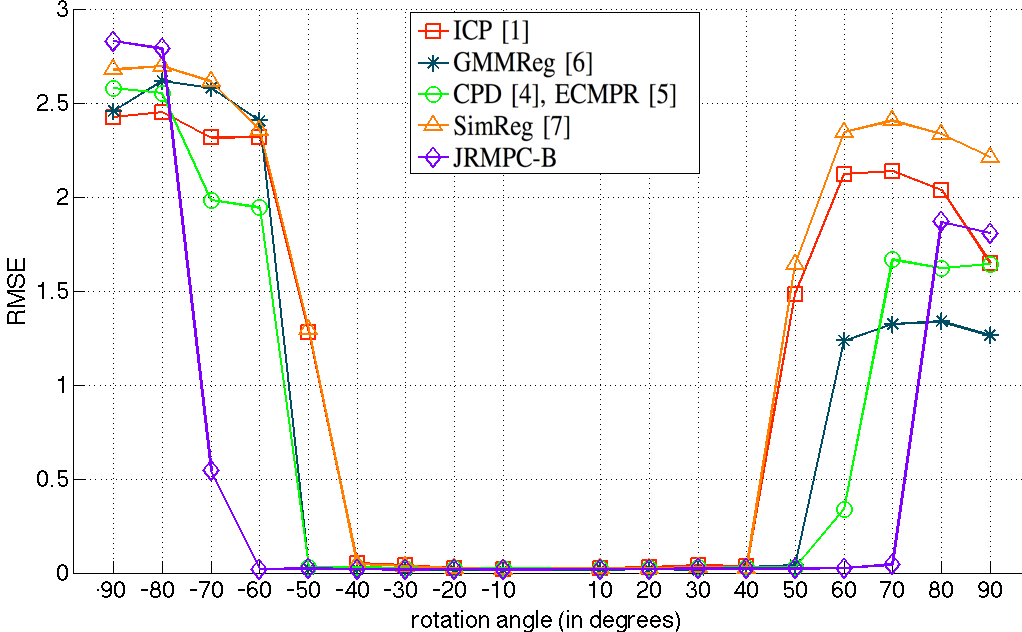}
&\includegraphics[width=.23\textwidth]{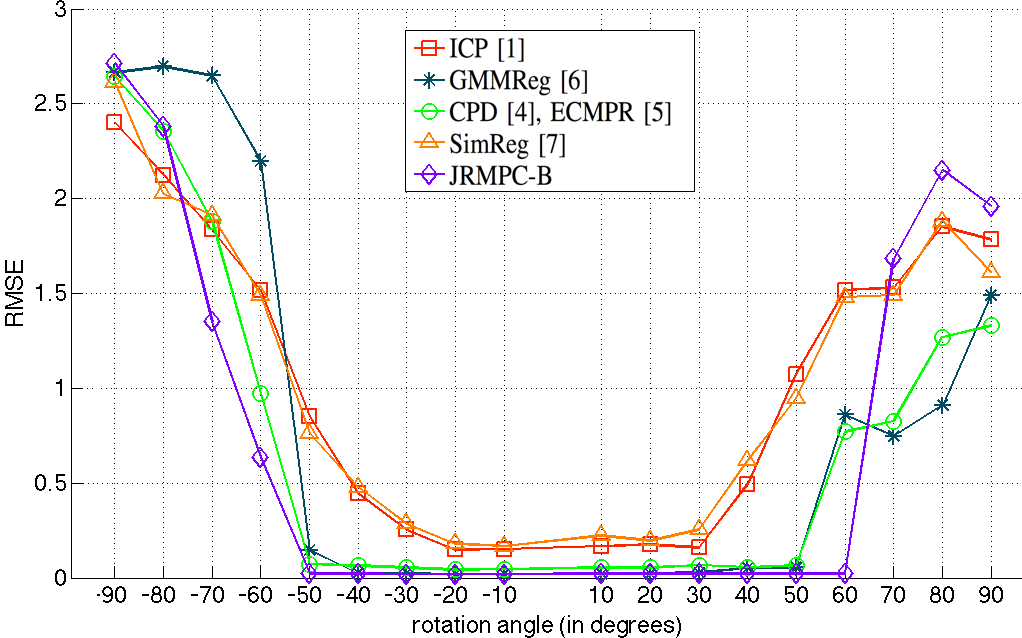}
&\includegraphics[width=.23\textwidth]{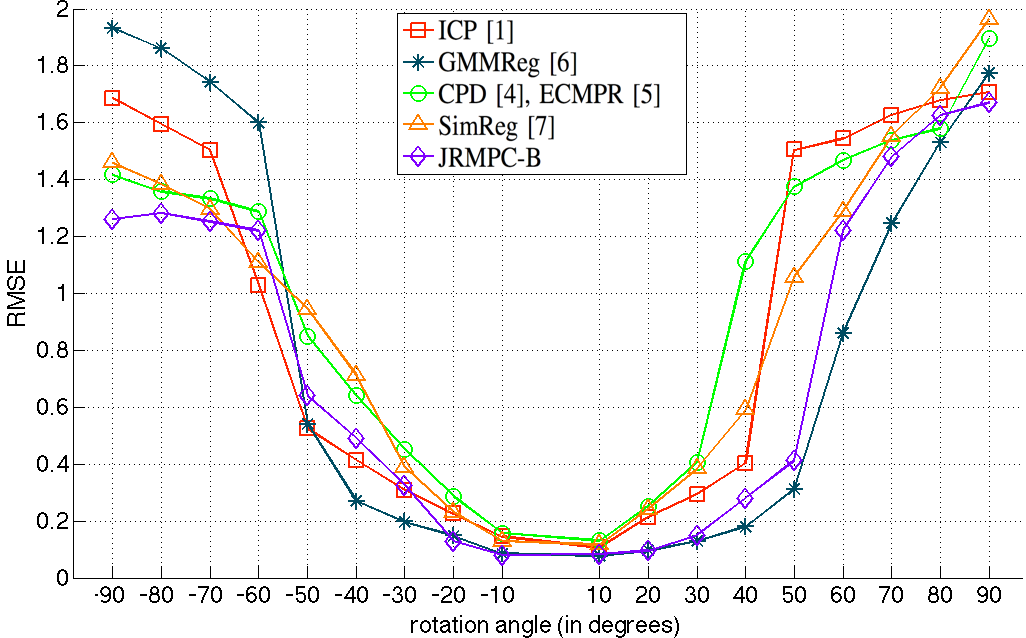}
&\includegraphics[width=.23\textwidth]{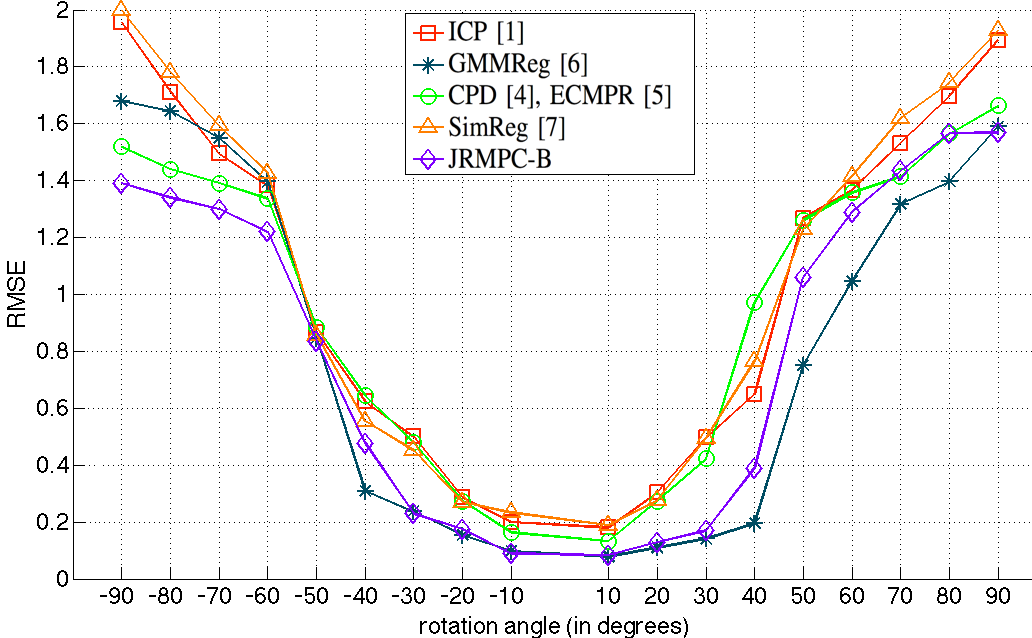}\\
(a) noise & (b) noise+outliers &(c) noise & (d) noise+outliers\\
\end{tabular}
\caption{RMSE as a function of the overlap (rotation angle) when two point sets are registered (SNR=$20dB$, $30\%$ outliers) (a),(b) ``Lucy'' (c), (d) ``Armadillo''}
\label{fig:two_view_experiment}
\end{figure*}
\section{Experiments}\label{sec:experiments}
In this section, we test and benchmark the proposed algorithms with widely used and publicly available 3D data, as well as with time-of-flight (TOF) and structured-light (Kinect) data. First, we compare the proposed algorithm with pairwise registration methods, which illustrates the behavior of the algorithm in comparison with other algorithms and in particular its robustness to noise and to outliers. Second, we compare our algorithm with recently proposed joint-registration algorithms. Third, we test and evaluate the best performing algorithms on challenging TOF data and on data captured with a moving Kinect sensor.
 
\subsection{Simulated Data}
\subsubsection{Comparison with pairwise registration algorithms}

We use 3D models from the Stanford 3D scanning repository\footnote{https://graphics.stanford.edu/data/3Dscanrep/}, i.e.,  ``Bunny'', ``Lucy'' and ``Armadillo", and we proceed as follows in order to synthesize multiple point sets from different viewpoints. The model is shifted around the origin, the points are downsampled and then rotated in the $xz$-plane; points with negative $z$ coordinates are rejected. This way, only a part of the object is viewed in each set, the point sets do not fully overlap, and the extent of the overlap depends on the rotation angle, as in real scenarios. It is important to note that downsampling differs over the sets, such that different points are present in each set as well as different cardinalities (from the range $[1000, 2000]$) are obtained. We add Gaussian noise to point coordinates based on a predefined signal-to-noise ratio (SNR), and  more importantly, we add outliers to each set which are uniformly distributed around five randomly chosen points of the set.  A tractable case of registering four point sets ($M=4$) is considered here, the angle between the first set and the other sets being $10^o$, $20^o$ and $30^o$ respectively. We include JRMPC-I in the latter experiments where more point-sets are registered.

\begin{table*}
\centering
\caption{Registration error of indirect mappings. For each model, the two first columns show the rotation error of $V_2\rightarrow V_3$ and $V_3\rightarrow V_4$ respectively, while the third column shows the standard deviation of the two errors ($SNR=10db$, $30\%$ outliers).}
\label{tab:proposed_vs_referenceBased}
\begin{tabular}{l|ccc|ccc|ccc}
\hline
~ & \multicolumn{3}{c|}{Bunny} & \multicolumn{3}{c|}{Lucy} & \multicolumn{3}{c}{Armadillo}\\ 
\hline
\hline
ICP~\cite{BeslMcKay92} & $0.329$ & $0.423$ & $0.047$ & $0.315$ & $0.297$ & $0.009$ & $0.263$ & $0.373$ & $0.055$ \\
GMMReg~\cite{JianVemuri2011} & $0.364$ & $0.303$ & $0.030$ & $0.129$ & $0.110$ & $0.009$ & $0.228$ & $0.167$ & $0.031$\\
CPD~\cite{MyronenkoPAMI2010}, ECMPR~\cite{Horaud2011} & $0.214$ & $0.242$ & $0.014$ & $0.144$ & $0.109$ & $0.017$ & $0.222$ & $0.204$ & $0.009$\\
SimReg~\cite{Williams2001} & $0.333$ & $0.415$ & $0.041$ & $0.354$ & $0.245$ & $0.055$ & $0.269$ & $0.301$ & $0.016$\\
JRMPC-B & $0.181$ & $0.165$ & $\bf{0.008}$ & $0.068$ & $0.060$ & $\bf{0.004}$ & $0.147$ & $0.147$ & $\bf{0.000}$\\
\hline
\end{tabular}
\end{table*}
For comparison, we consider the following baselines that follow the one-vs-all approach: ICP~\cite{BeslMcKay92}, CPD~\cite{MyronenkoPAMI2010}, ECMPR~\cite{Horaud2011}, GMMReg~\cite{JianVemuri2011}. In addition, we include a sequential version of ICP (seqICP) and a modification of~\cite{Williams2001}, abbreviated here as SimReg. Unlike the original version, the latter allows updating the matches at each iteration. Recall that CPD is exactly equivalent to ECMPR when it comes to rigid registration.\footnote{CPD considers common variance for all components, while each component has its own variance with ECMPR; that latter case is considered here.} As showed in~\cite{JianVemuri2011}, Levenberg-Marquardt ICP~\cite{Fitzgibbon2003} performs similarly with GMMReg, while~\cite{Wang2008} shows that GMMReg is superior to Kernel Correlation~\cite{TsinKanade2004}. As a consequence, we implicitly assume a variety of baselines. All the competitors employ $M-1$ registrations between the first and rest sets, while SimReg considers all the pairs of (overlapping) sets.

To evaluate the performance. we use the root of the mean squared error (RMSE) of the rotation parameters averaged by the number of sets. For all algorithms, we implicitly initialise the translations by transferring the centroids of the point clouds into the same point, while identity matrices initialize the rotations. GMMReg and SimReg are kind of favored in the comparison, since the former benefits from a two-level optimization (the first level initializes the second one) while the latter starts from the point where the pairwise ICP ends. Notice that the proposed method provides a transformation for \emph{every} point set, while ground rotations are typically expressed in terms of the first set. Hence, the product of estimations $\hat{\bv{R}}_{1}^\top\hat{\bv{R}}_j$ is compared with the ground rotation $\bv{R}_j$, i.e., the error for the $j$-th set is $\|\hat{\bv{R}}_{1}^\top\hat{\bv{R}}_j-\bv{R}_j\|_F$.

JRMPC starts here from a completely unknown GMM where the initial means $\boldsymbol{\mu}_k$ are distributed on a sphere that spans the convex hull of the sets. The variances $\sigma_k$ are here initialized with the median distance between $\boldsymbol{\mu}_k$ and all the points in $\Vmat$. We found that updating the priors does not drastically improve the registration. We therefore keep them constant and equal to $1/(K+1)$ during EM, while $h$ is chosen to be the volume of a sphere whose radius is $0.5$; the latter is not an arbitrary choice because the point coordinates are normalized by the maximum distance between points of the convex hull of $\Vmat$.
The number of the components, $K$, is here equal to $60\%$ of the mean cardinality. We use $100$ iterations for all algorithms while GMMReg performs $10$ and $100$ function evaluations for the first and second optimization levels respectively. However, the current authors' implementation allows to extract the parameters after the latest evaluation.

\begin{figure*}
\begin{tabular}{ccc}
\includegraphics[width=.31\textwidth]{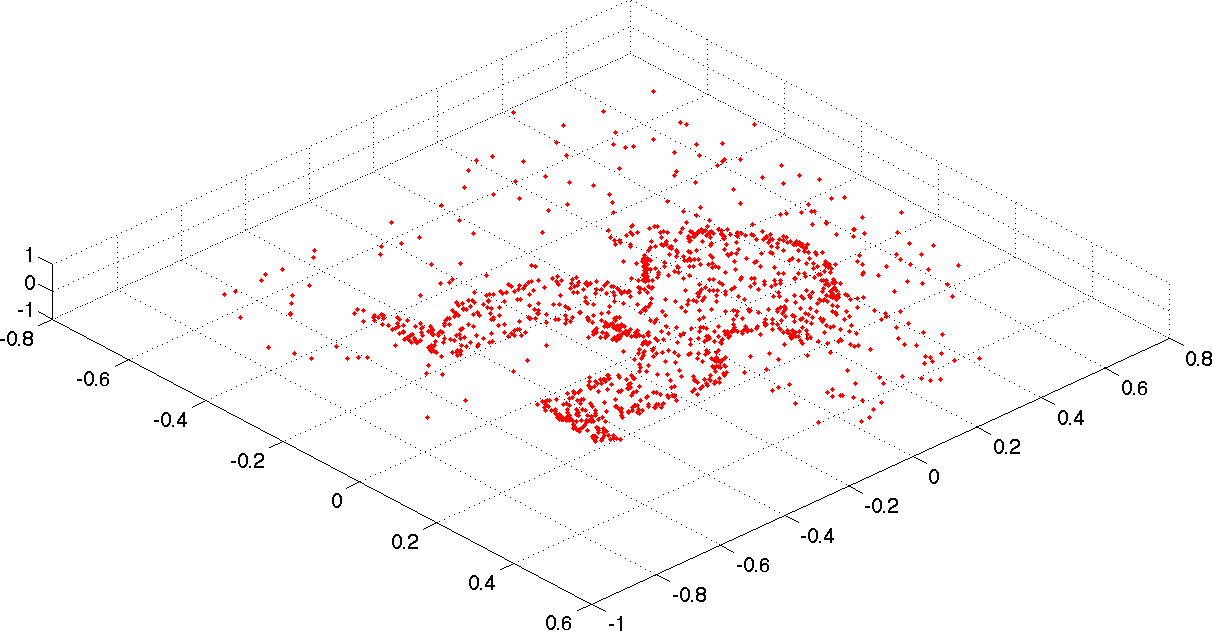}
&\includegraphics[width=.31\textwidth]{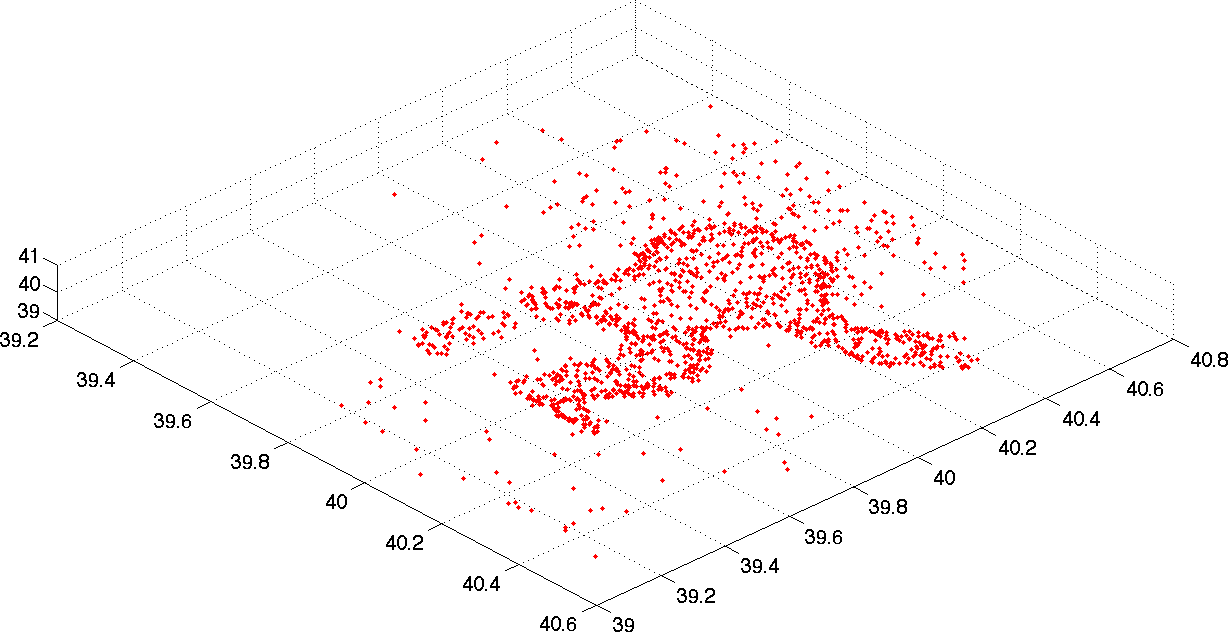}
&\includegraphics[width=.31\textwidth]{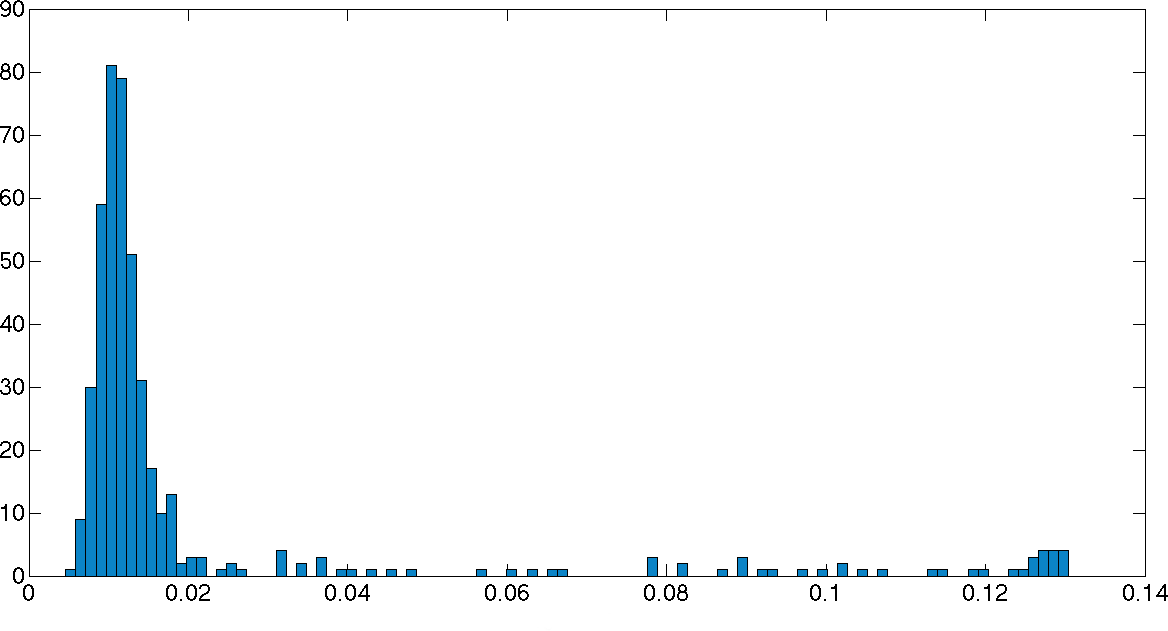}\\
(a) & (b) & (c)\\
\includegraphics[width=.31\textwidth]{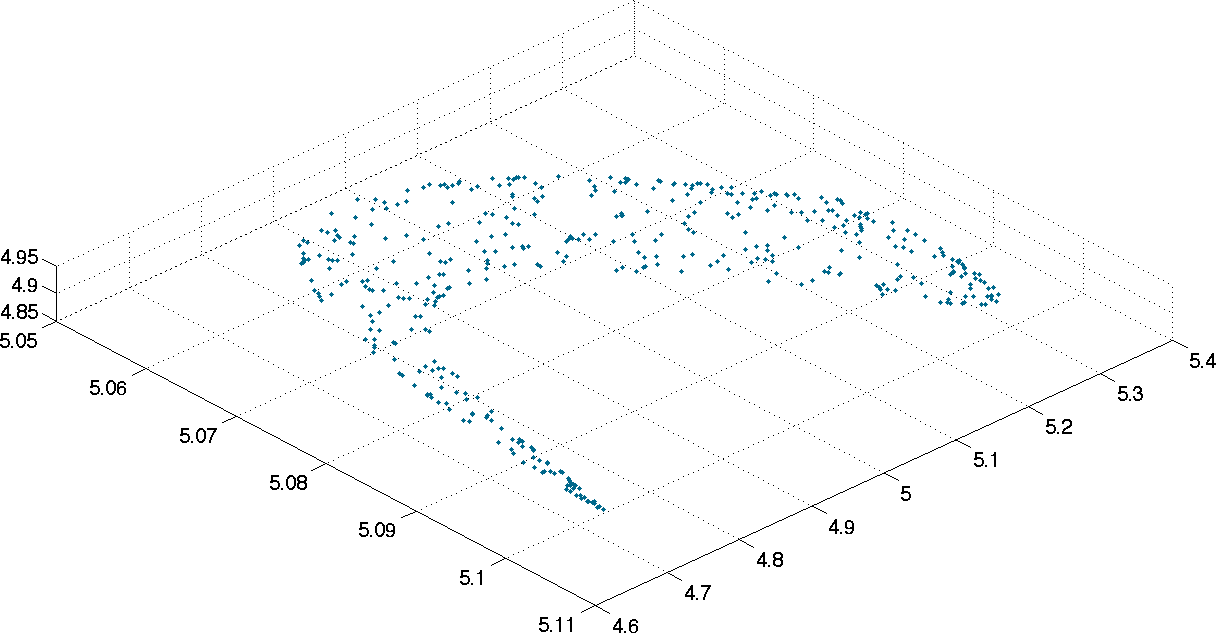}
&\includegraphics[width=.31\textwidth]{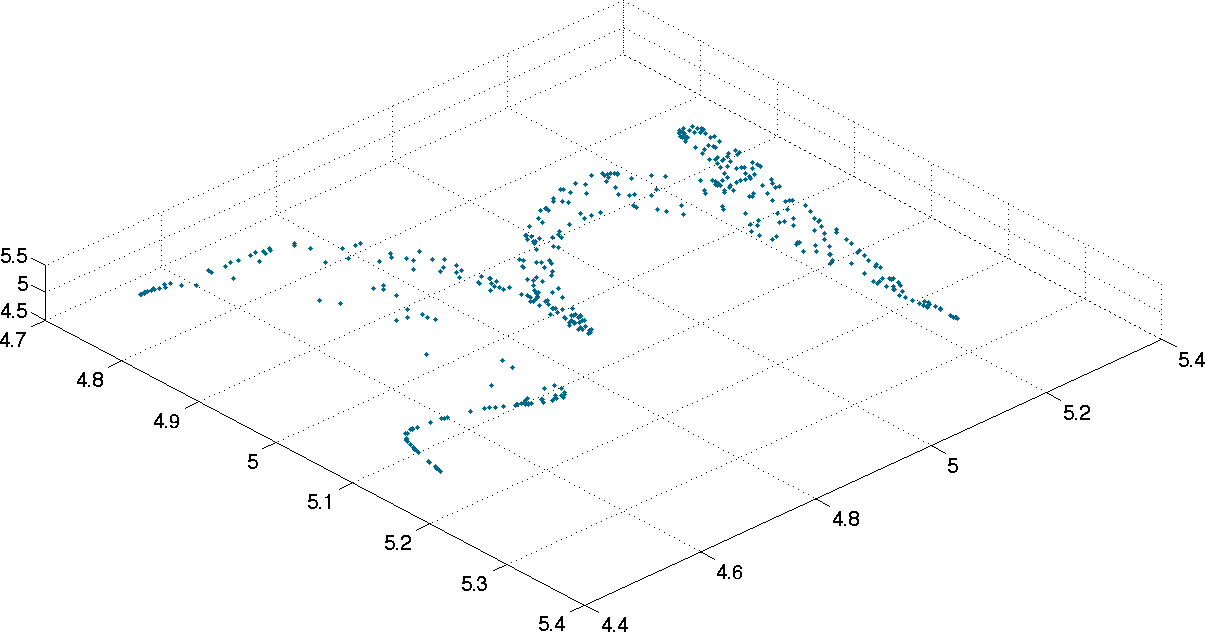}
&\includegraphics[width=.31\textwidth]{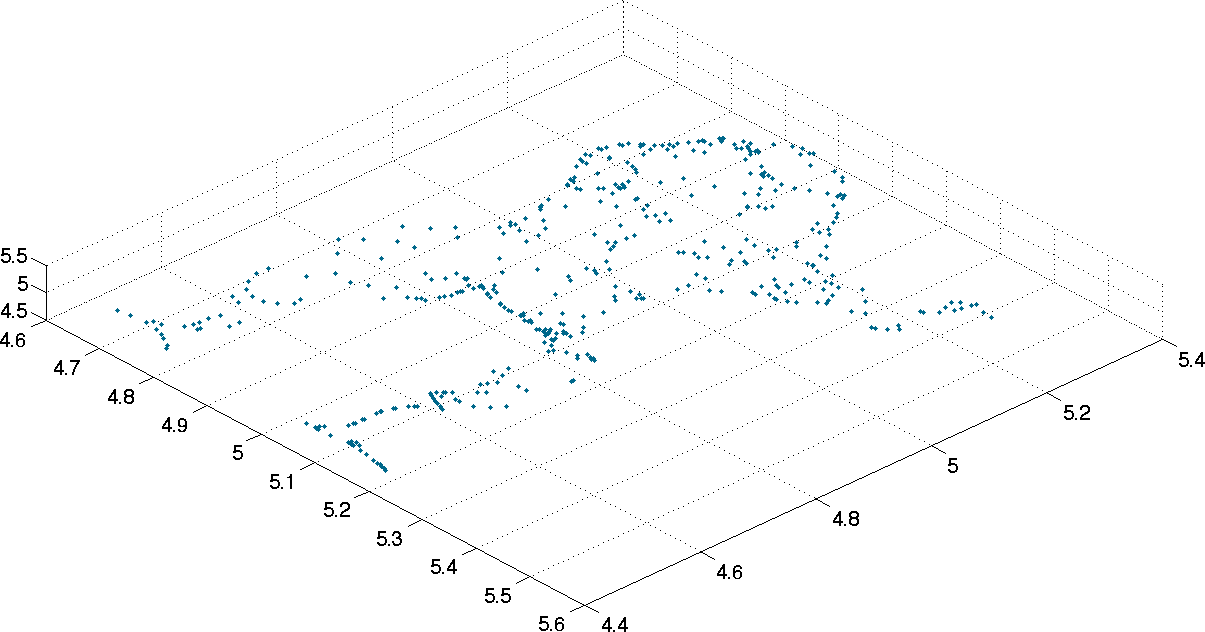}\\
(d) & (e) & (f)\\
\includegraphics[width=.31\textwidth]{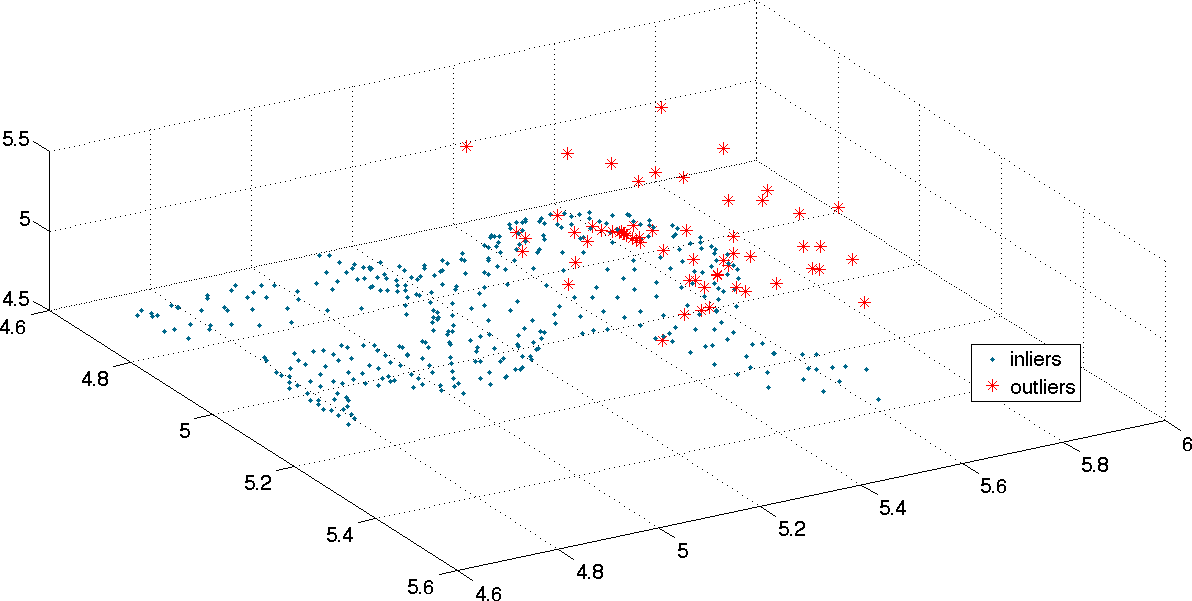}
& \includegraphics[width=.31\textwidth]{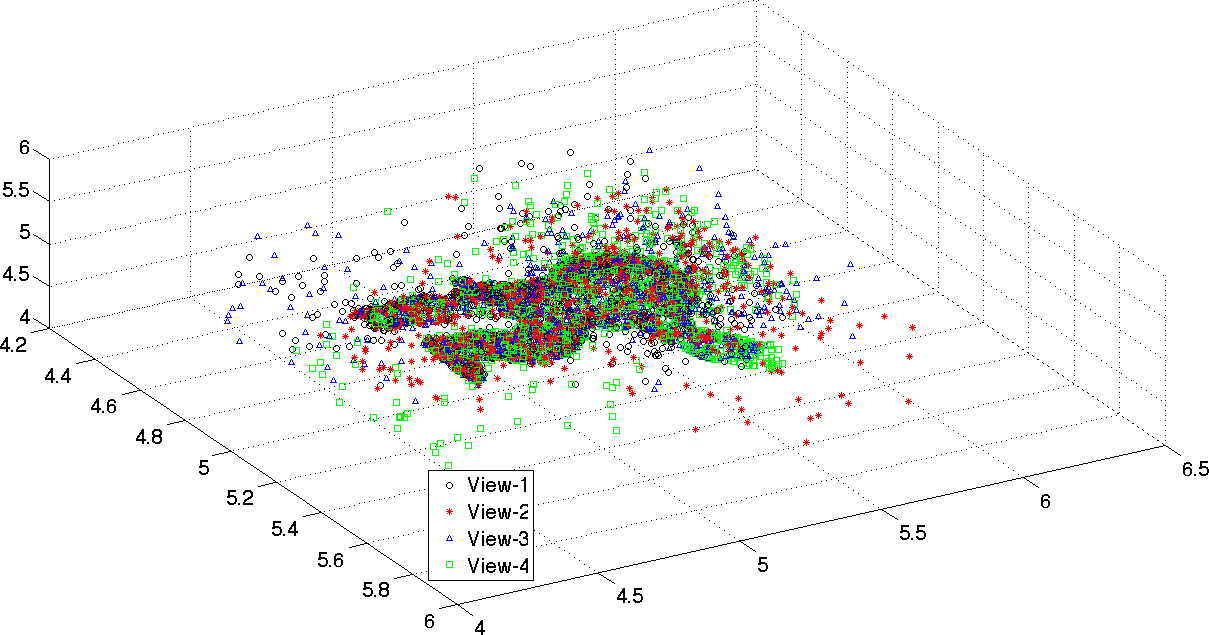}
& \includegraphics[width=.31\textwidth]{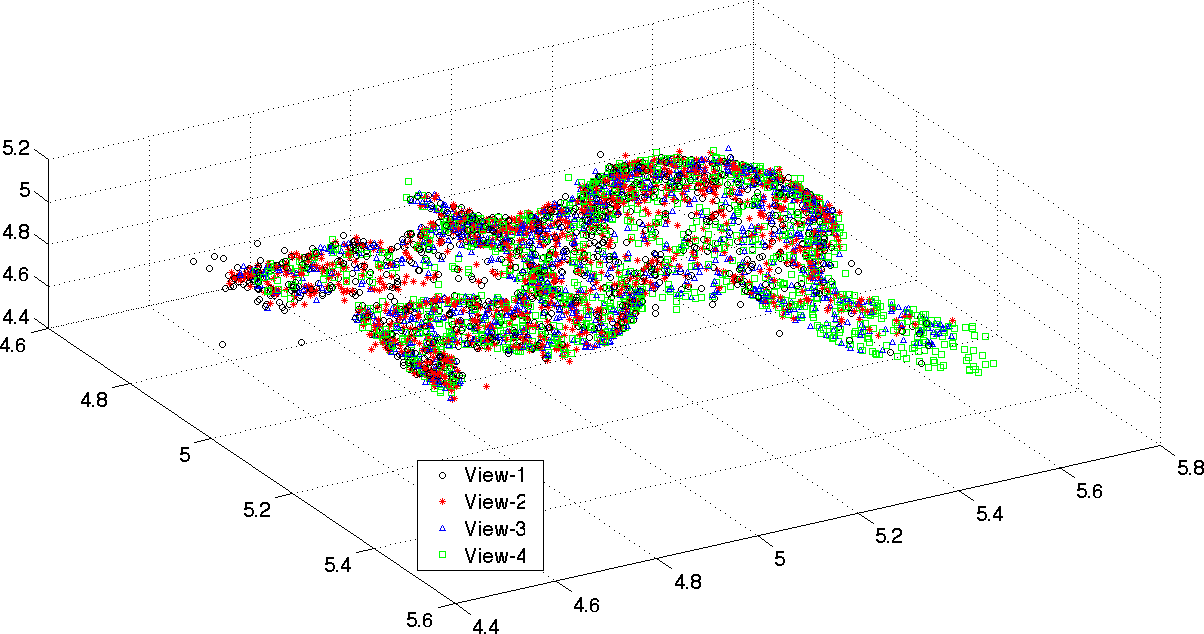}\\
(g) & (h) & (i)\\
\end{tabular}
\caption{ (a),(b) Two point sets (out of four) with outliers; (c) distribution of estimated variances; instances of GMM means after (d) $5$, (e) $15$, and (f) $30$ iterations; (g) the splitting of model points into inliers and outliers; joint-registration of four point sets (h) before and (i) after removing ``bad'' points (\emph{best viewed on-screen}).}
\label{fig:mean_shape}
\end{figure*}

Fig.~\ref{fig:dragon_noise_and_outliers} shows the final log-RMSE, averaged over $100$ realisations and all views, as a function of outlier percentage for each 3D model. Apparently, ICP and SimReg are more affected by the presence of outliers owing to one-to-one correspondences. CPD and GMMReg are affected in the sense that the former assigns outliers to any of the GMM components, while the latter may merge outliers into clusters. The proposed method is more robust to outliers and the registration is successful even with densely present outliers. The behavior of the proposed algorithm in terms of the outliers is discussed in detail below and showed on Fig.~\ref{fig:mean_shape}. To visualize the convergence rate of the algorithms, we show curves for a challenging setting ($SNR=10dB$ and $20\%$ outliers). Regarding GMMReg, we just plot a line that shows the error in steady state, since the author's implementation allow to extract the final parameters only. There is a performance variation as the model's surface changes. ``Lucy'' is more asymmetric than ``Bunny'' and ``Armadillo'', thus a lower floor is achieved. Unlike the competitors, JRMPC-B may show a minor perturbation in the first iterations owing to the joint solution and the initialization of the means and the variances. 

It is also important to show the estimation error between sets whose geometric relation is not directly estimated. This also shows how biased each algorithm is. Based on the above experiment (SNR=$10db$, $20\%$ outliers), Table~\ref{tab:proposed_vs_referenceBased} reports the average rotation error for the pairs $(V_2,V_3)$ and $(V_3,V_4)$, as well as the standard deviation of these two errors as a measure of bias. All but seqICP do not estimate these individual mappings alone. 
The proposed scheme, not only provides the lowest error, but it also offers the most symmetric solution. 

A second experiment evaluates the robustness of the algorithms in terms of the rotation angle between two point sets, that is, the extent of their overlap. This also allows us to show how the proposed algorithm deals with the simple case of two point sets. Recall that JRMPC-B does not reduce to CPD/ECMPR in the two-set case, but it still computes the poses of the two sets with respect to the ``central" GMM. Fig.~\ref{fig:two_view_experiment} plots the average RMSE over $50$ realizations of ''Lucy`` and ``Armadillo'',  when the relative rotation angle varies from $-90^o$ to $90^o$.
As for an acceptable registration error, the proposed scheme achieves the widest and shallowest basin for ``Lucy'', and  competes GMMReg for ``Armadillo''. Since ``Armadillo'' consists of smooth and concave surfaces, the performance of the proposed scheme is better with multiple point sets than the two-set case, hence the difference with GMMReg. The wide basin of GMMReg is also due to its sophisticated initialization.  

As mentioned, a by-product of the proposed method is the reconstruction of an outlier-free model. In addition, we are able to detect the majority of the outlying points based on the variance of the component they most likely belong to. To show this effect, we use the results of one realization of the first experiment with $30\%$ outliers. Fig.~\ref{fig:mean_shape} shows in (a) and (b) two out of four point sets, thereby one verifies  the distortion of the point sets, as well as how different the sets may be, e.g., the right hand is missing in the first set. The progress of $\boldsymbol{\mu}_k$ estimation is shown in (d-f). Apparently, the algorithm starts by reconstructing the scene model (observe the presence of the right hand). Notice the size increment of the hull of the points $\boldsymbol{\mu}_k$, during the progress. This is because the posteriors in the first iteration are very low and make the means $\boldsymbol{\mu}_k$ shrink into a very small cell. While the two point sets are around the points $(0,0,0)$ and $(40,40,40)$, we build the scene model around the point $(5,5,5)$. The distribution of the final deviations $\sigma_k$ is shown in (c). We get the same distribution with any model and any outlier percentage, as well as when registering real data. Although one can fit a pdf, e.g., Rayleigh, it is convenient enough here to split the components using the threshold $T_{\sigma} = 2\times{median}(\mathcal{S})$, where $\mathcal{S}=\{\sigma_k| k=1,\ldots K\}$.  Accordingly, we build the scene model and we visualize the binary classification of points $\boldsymbol{\mu}_k$. Apparently, whenever components attract outliers, even not far from the object surface, they tend to spread their hull by increasing their scale.  Based on the above thresholding, we can detect such components and reject points that are assigned with high probability to them, as shown in (g). Despite the introduction of the uniform component that prevents the algorithm from building clusters away from the object surface, locally dense outliers are likely to create components outside the surface. In this example, most of the point sets contain outliers above the shoulders, and the algorithm builds components with outliers only, that are post-detected by their variance. The integrated surface is shown in (h) and (i) when ``bad" points are automatically removed. Of course, the surface can be post-processed, e.g., smoothing, for a more accurate representation, but this is beyond of our goal. 



\begin{figure*}
\begin{center}
\begin{tabular} {cccc}
\includegraphics[width=0.2\textwidth]{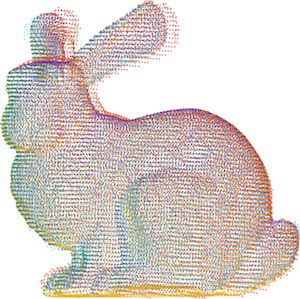} 
& \includegraphics[width=0.2\textwidth]{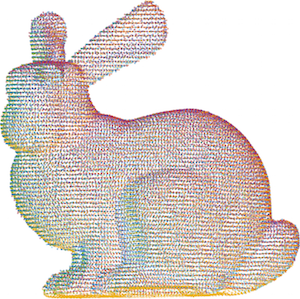}
& \includegraphics[width=0.2\textwidth]{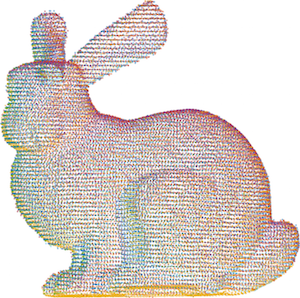}
& \includegraphics[width=0.2\textwidth]{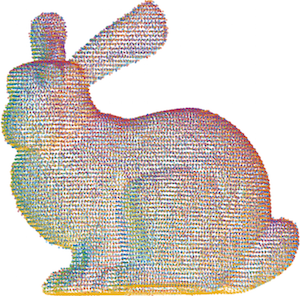}\\
\includegraphics[width=0.2\textwidth]{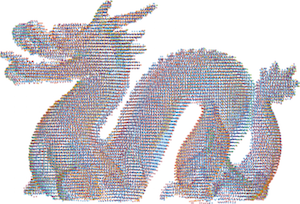} 
& \includegraphics[width=0.2\textwidth]{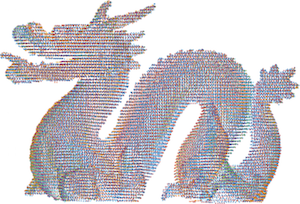}
& \includegraphics[width=0.2\textwidth]{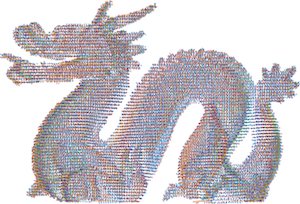}
& \includegraphics[width=0.2\textwidth]{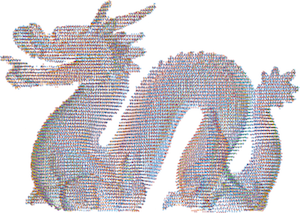}\\
\includegraphics[width=0.1\textwidth]{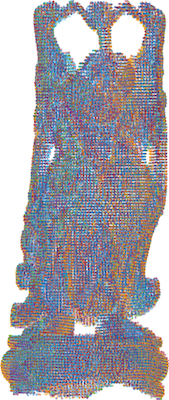} 
& \includegraphics[width=0.1\textwidth]{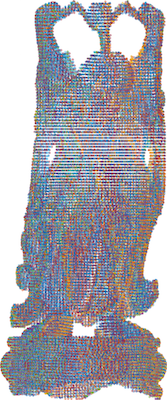}
& \includegraphics[width=0.1\textwidth]{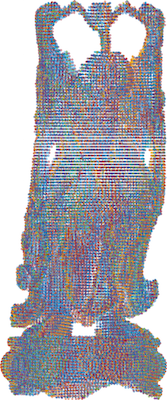}
& \includegraphics[width=0.1\textwidth]{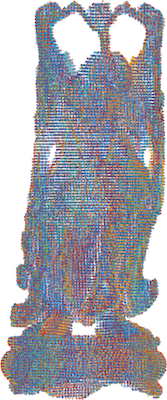}\\
(a) MAICP & (b) MATrICP & (c) JRMPC-B & (d) JRMPC-I 
\end{tabular}
\end{center}
\caption{Integrated models of Bunny (first row), Dragon (second row) and Happy Buddha (third row) based on four joint-wise registration methods (\emph{best viewed on-screen}).}
\label{fig:reconstructions}
\end{figure*}

\subsubsection{Comparison with joint registration algorithms}
We here compare our method with the joint registration algorithms of \cite{Govindu2014} and \cite{ZhongyuLi2014}.\footnote{The code was kindly provided by the authors.} Recall that both rely on the motion averaging strategy using the ICP and the trimmed-ICP algorithm, respectively, hence abbreviated as MAICP and MATrICP. According to the literature, MAICP and MATrICP seem to outperform the methods of~\cite{Sharp2004,Benjemaa1998,Williams2001,Bergevin1996}. The method of~\cite{Williams2001} is also included here as a baseline that considers fixed matches between the sets, and is referred to as multi-view ICP (MV-ICP). As mentioned above, MV-ICP considers all the pairs of overlapping views. While~\cite{Wang2008} generalizes GMMReg~\cite{JianVemuri2011} for multiple point-sets, the authors provide the code for two-set case only.

\begin{table*}
\centering
\caption{Comparison of multi-view registration methods without adding noise}
\label{tab:multiview-reg-results}
\begin{tabular}{lccccccc}
\hline
~ & Raw-data & Initialization & MV-ICP~\cite{Williams2001} & MAICP~\cite{Govindu2014} & MATrICP~\cite{ZhongyuLi2014} & JRMPC-B & JRMPC-I \\
\hline
\hline
Bunny & 3.45 & 2.10 &1.54 & 0.95 & \bf{0.27} & 0.37 & 0.69\\
Dragon & 7.28 & 4.37 & 3.75 & 1.95 & 0.62 & \bf{0.47} & 0.73\\
Happy Buddha & 10.77 & 3.18 & 2.45 & 0.64 & 0.43 & \bf{0.36} & 0.77\\
\hline
\end{tabular}
\end{table*}

For consistency reasons, the experimental setup of \cite{Govindu2014} is adopted, i.e., the point-sets have been roughly pre-aligned using a standard pair-wise ICP scheme. The error metric is the angle (in degrees) obtained from the composition of true and inverse estimation averaged over all point-sets, that is, it should ideally vanish.  As with \cite{Govindu2014}, we use the ``Bunny'', ``Dragon'' and ``Happy Buddha''  models from Stanford scanning repository owing to the availability of the ground truth motions. While ``Bunny'' is asymmetrically captured from $10$ viewpoints, the last two sets contain $15$ scans from evenly spaced view angles (every $24$ degrees). To get the point-sets, true transformations first apply to the sets and then, we deform each set by a random yet known transformation. Finally, we down-sample the point-sets so that the cardinalities vary from $2000$ to $5000$ points. Unlike \cite{Govindu2014} and \cite{ZhongyuLi2014}, we also evaluate the registration performance, when the point-sets have been further perturbed by noise. We deliberately avoid adding outliers since any mis-registration in the initialization step would make the motion averaging methods completely fail. 
\begin{figure*}
\begin{center}
\begin{tabular} {cccccc}
\includegraphics[height=1.5cm]{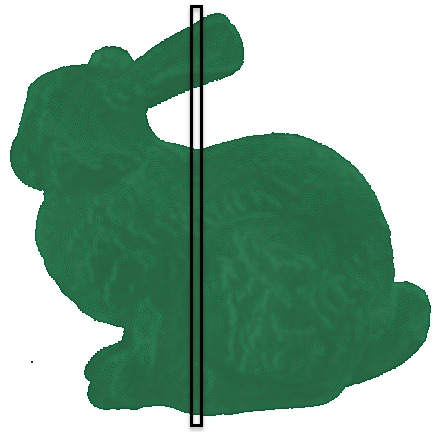} 
& \includegraphics[width=0.15\textwidth]{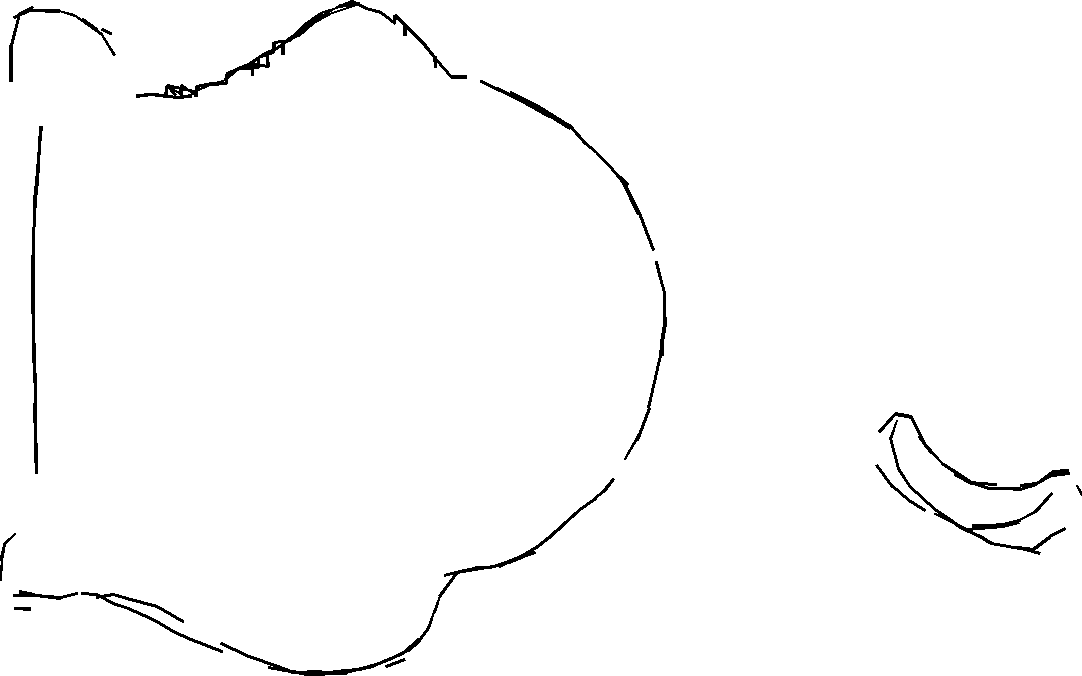}
& \includegraphics[width=0.15\textwidth]{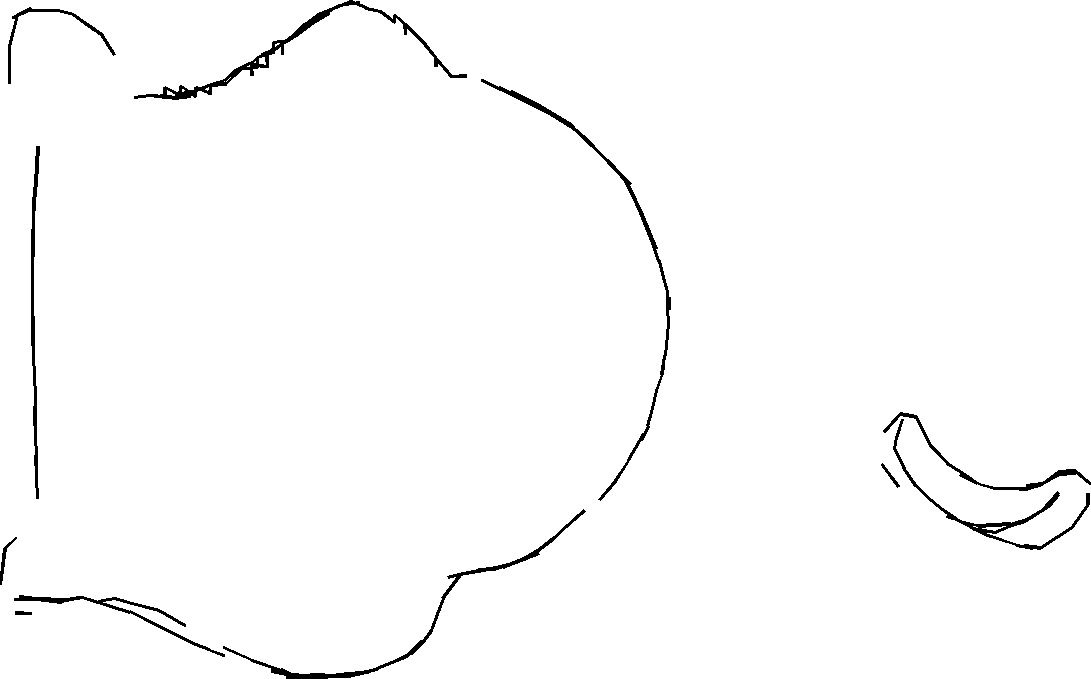}
& \includegraphics[width=0.15\textwidth]{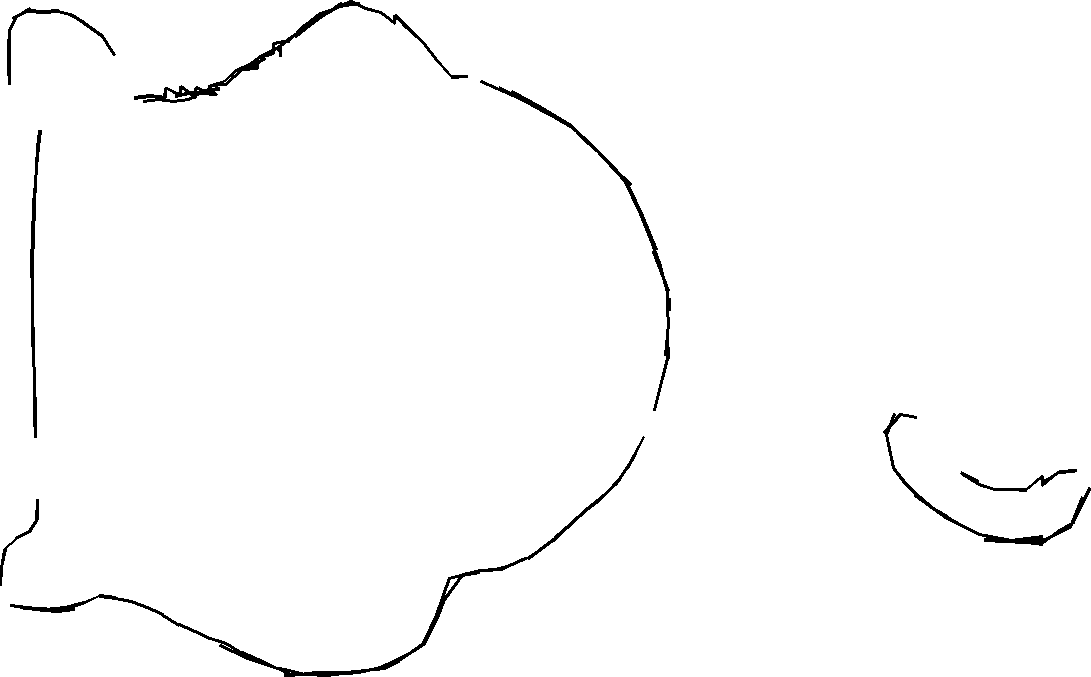}
& \includegraphics[width=0.15\textwidth]{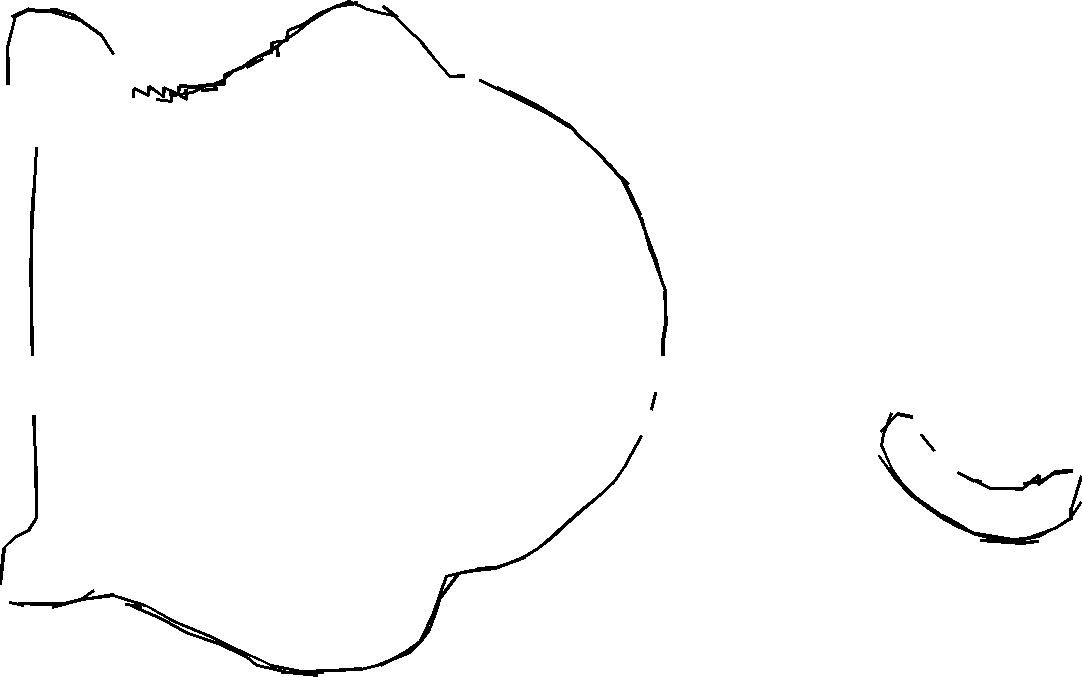}
& \includegraphics[width=0.15\textwidth]{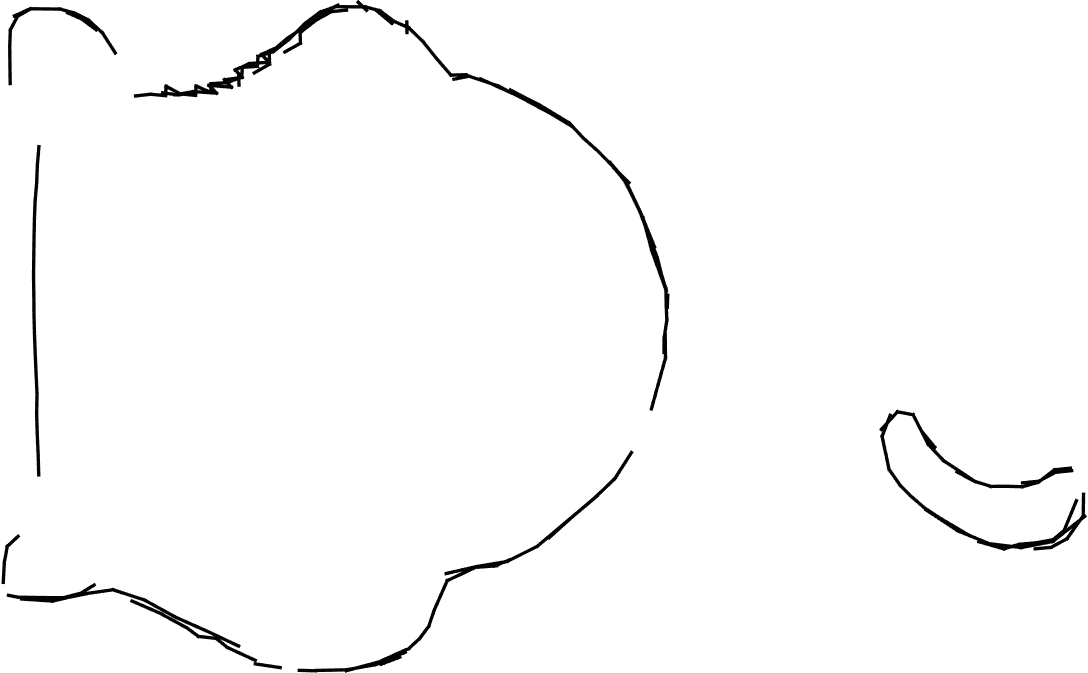}\\
\includegraphics[height=1.3cm]{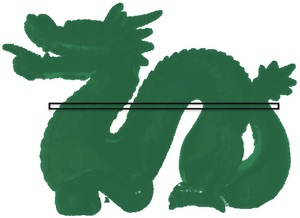} 
& \includegraphics[width=0.15\textwidth]{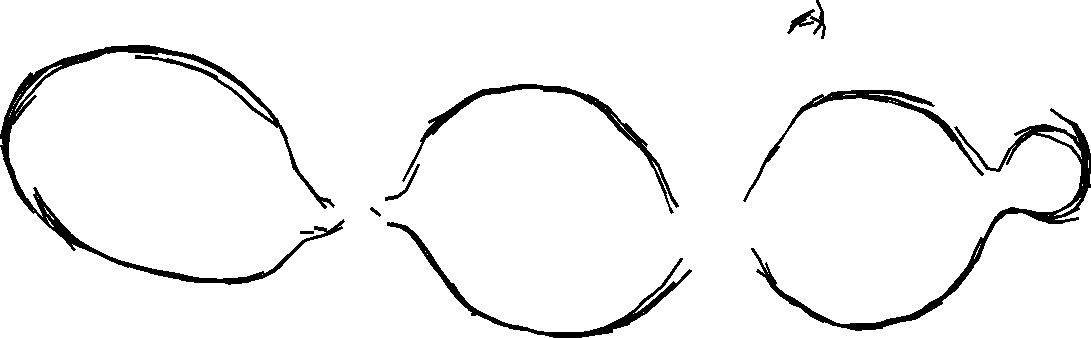}
& \includegraphics[width=0.15\textwidth]{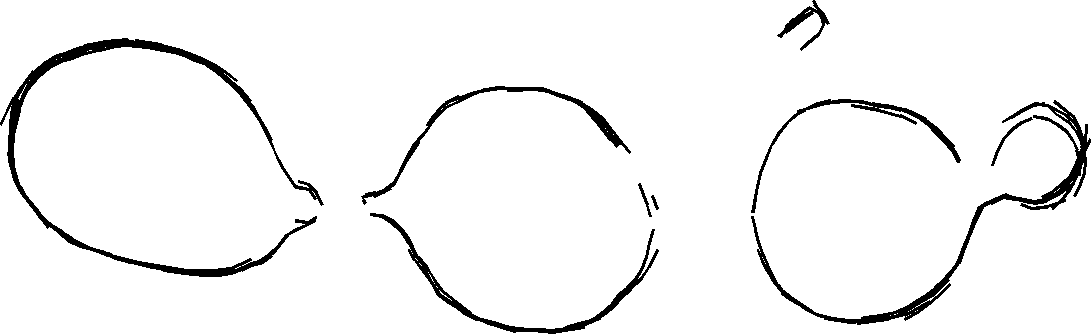}
& \includegraphics[width=0.15\textwidth]{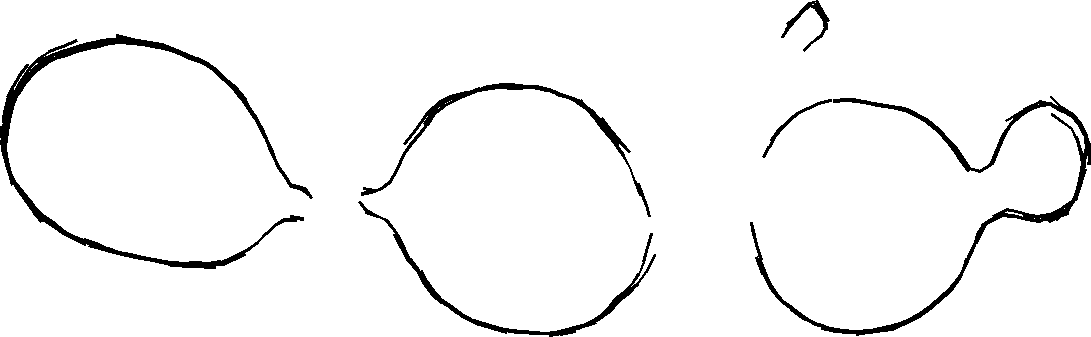}
& \includegraphics[width=0.15\textwidth]{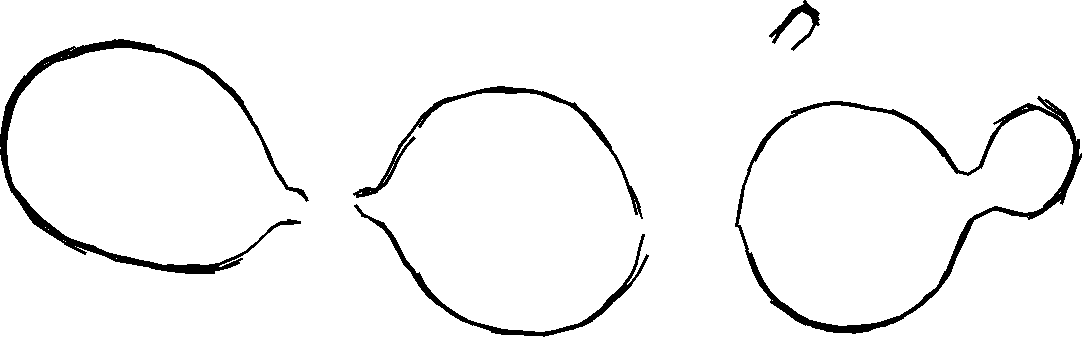}
& \includegraphics[width=0.15\textwidth]{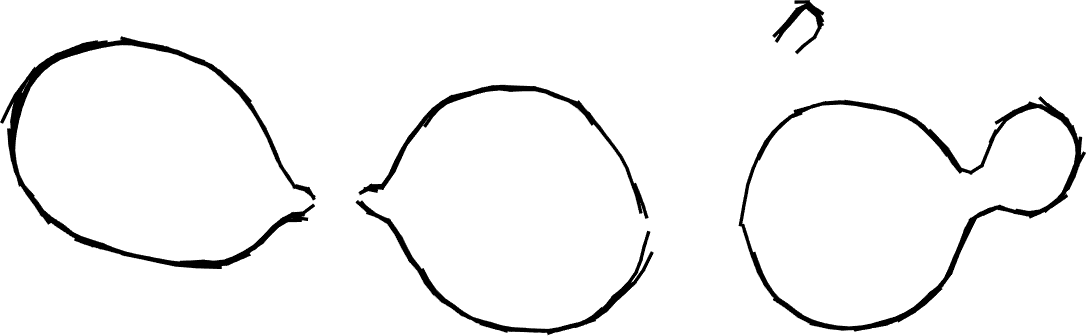}\\
\includegraphics[height=2.0cm]{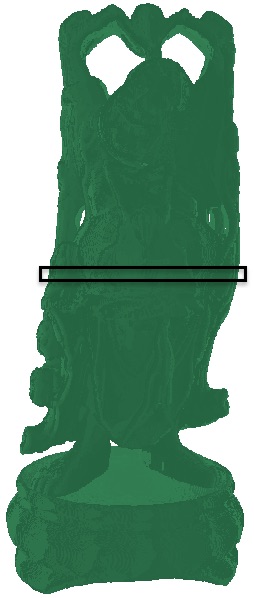} 
& \includegraphics[width=0.1\textwidth]{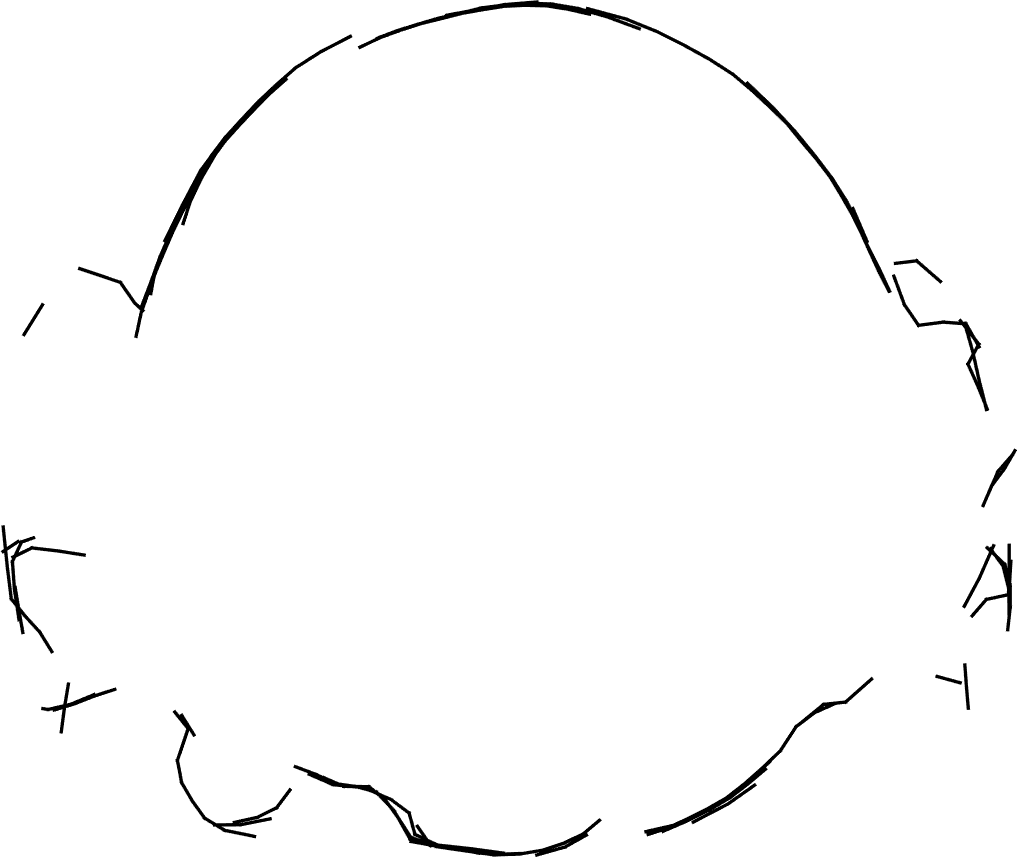}
& \includegraphics[width=0.1\textwidth]{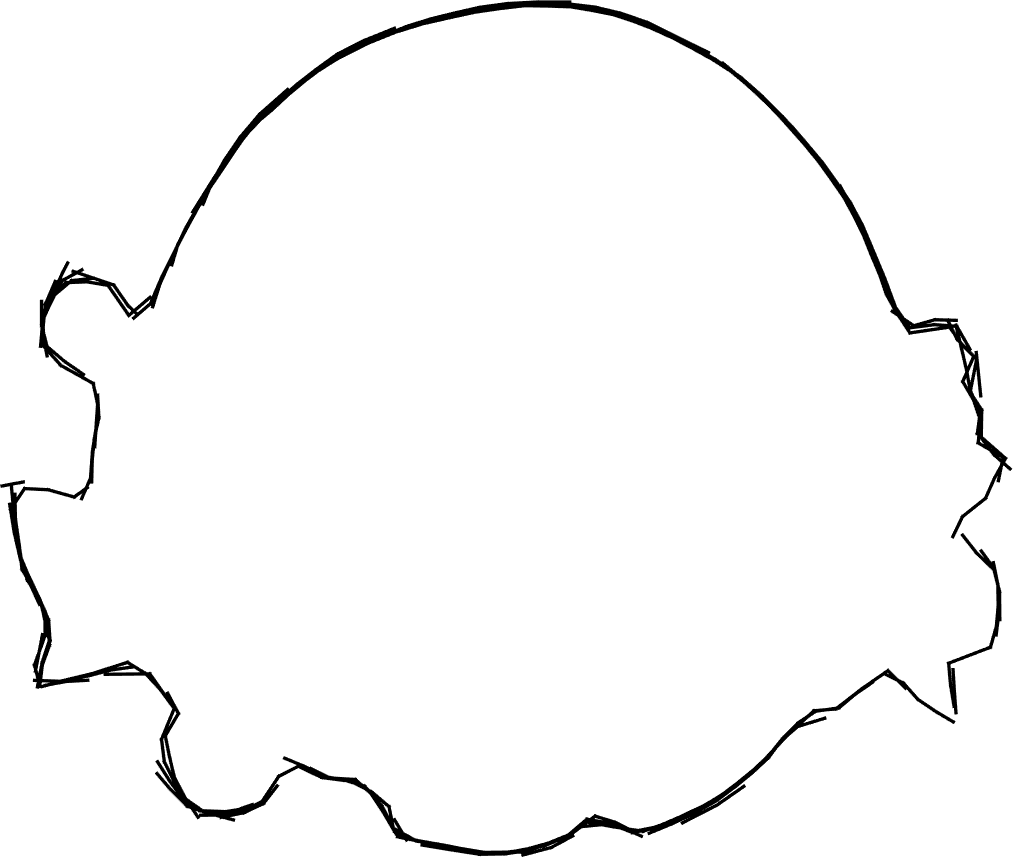}
& \includegraphics[width=0.1\textwidth]{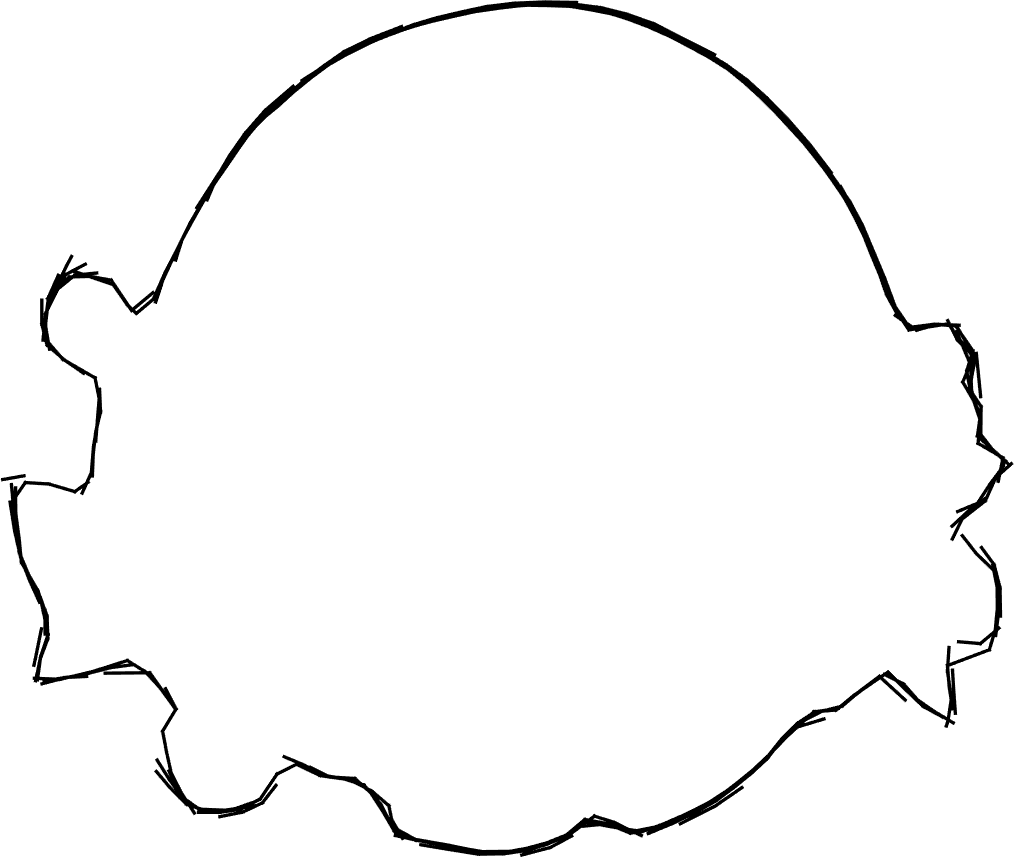}
& \includegraphics[width=0.1\textwidth]{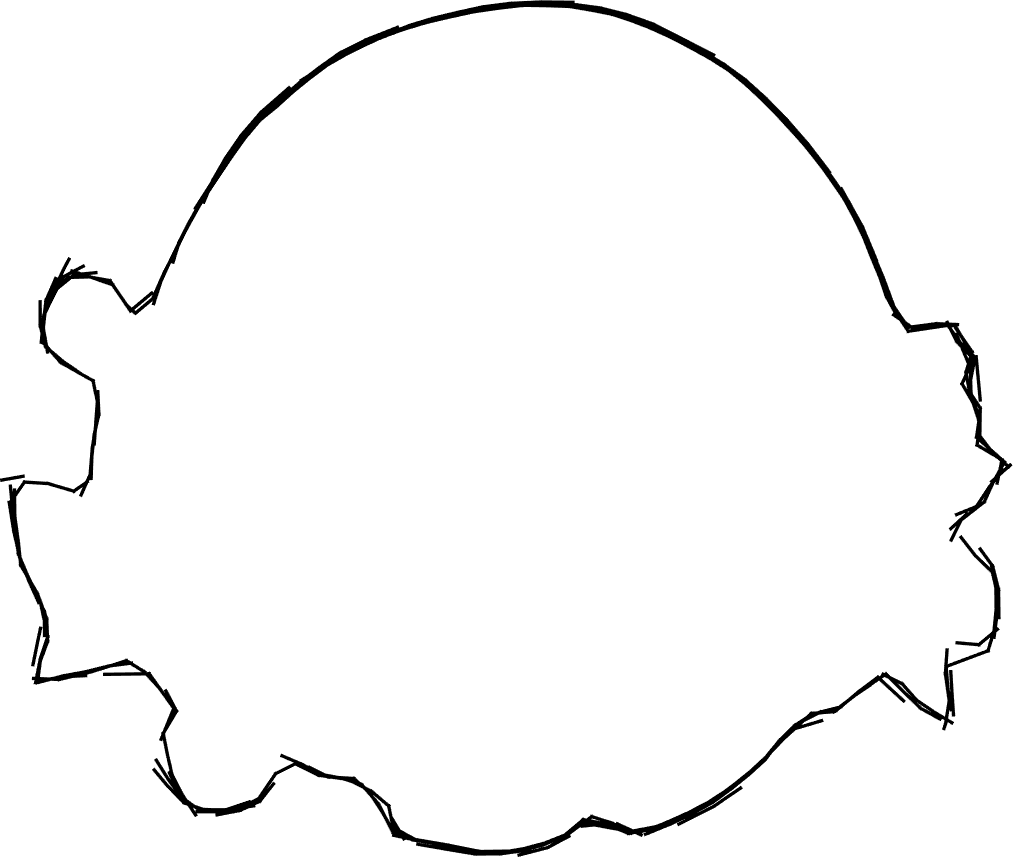}
& \includegraphics[width=0.1\textwidth]{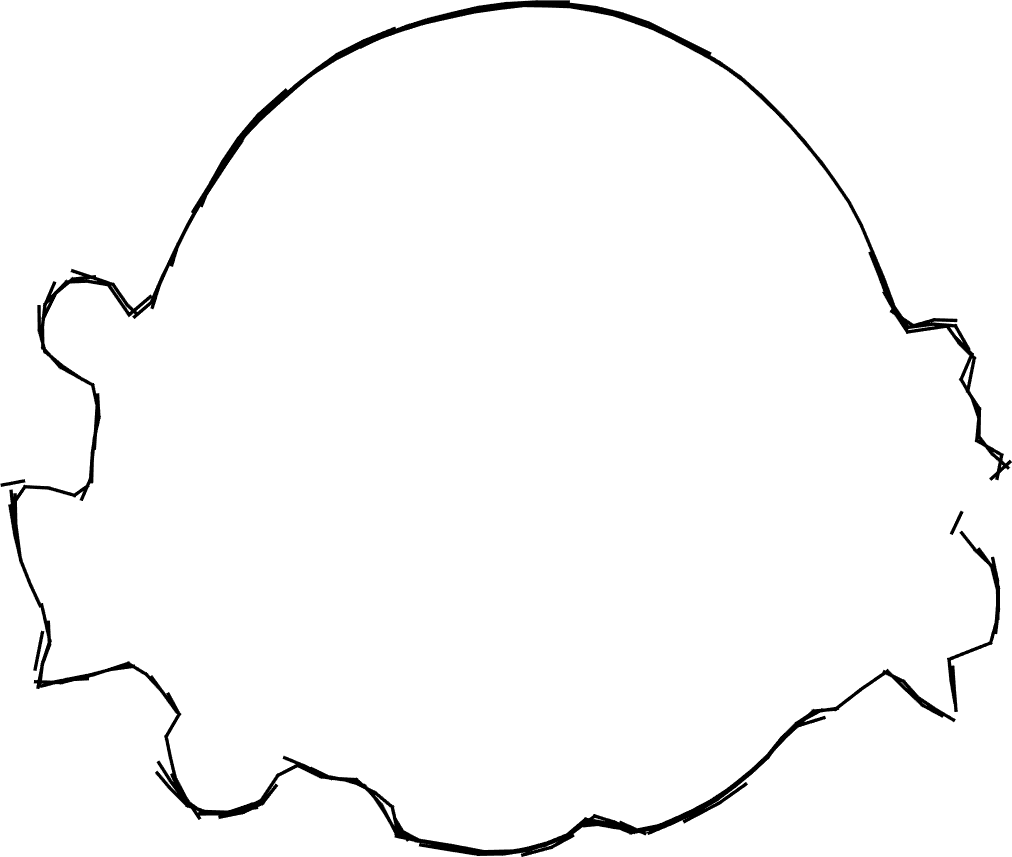}\\
cross-section area & pairwise ICP & MAICP & MATrICP & JRMPC-B & JRMPC-I
\end{tabular}
\end{center}
\caption{Cross-section of Bunny (top), Dragon (middle) and Happy Buddha (bottom) obtained from several algorithms (\emph{best viewed on-screen}).}
\label{fig:cross-sections}
\end{figure*}\begin{figure}
\begin{center}
\begin{tabular} {cc}
\includegraphics[width=0.42\columnwidth]{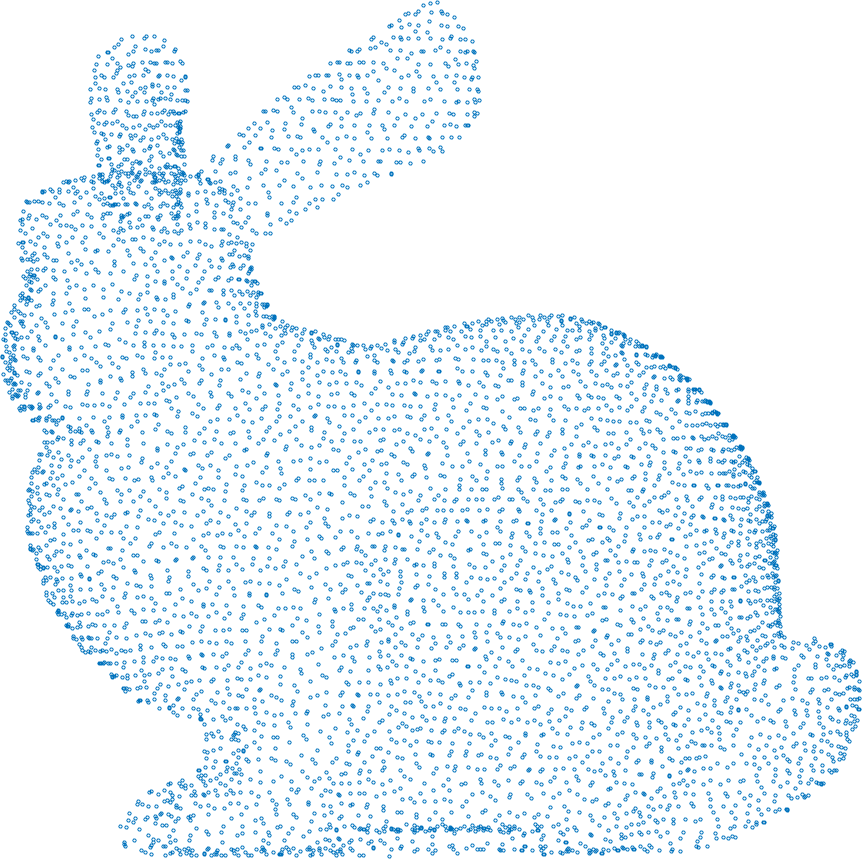}
& \includegraphics[width=0.42\columnwidth]{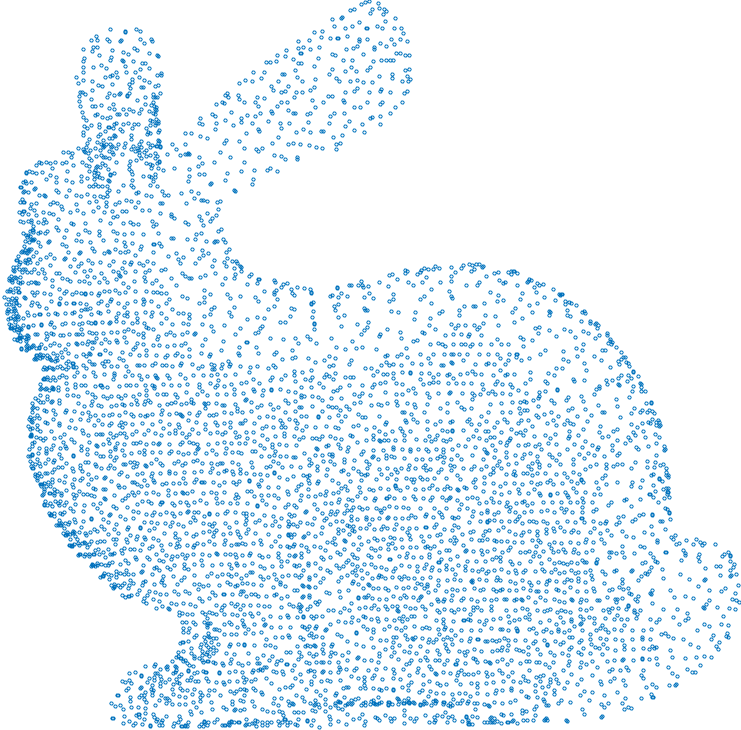}\\
\includegraphics[width=0.42\columnwidth]{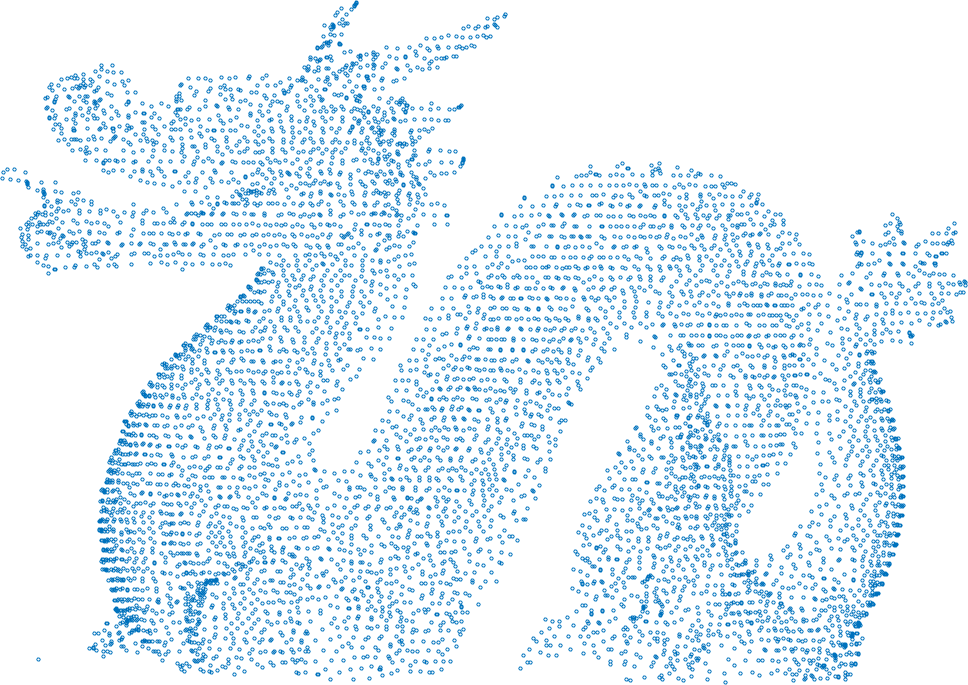}
& \includegraphics[width=0.42\columnwidth]{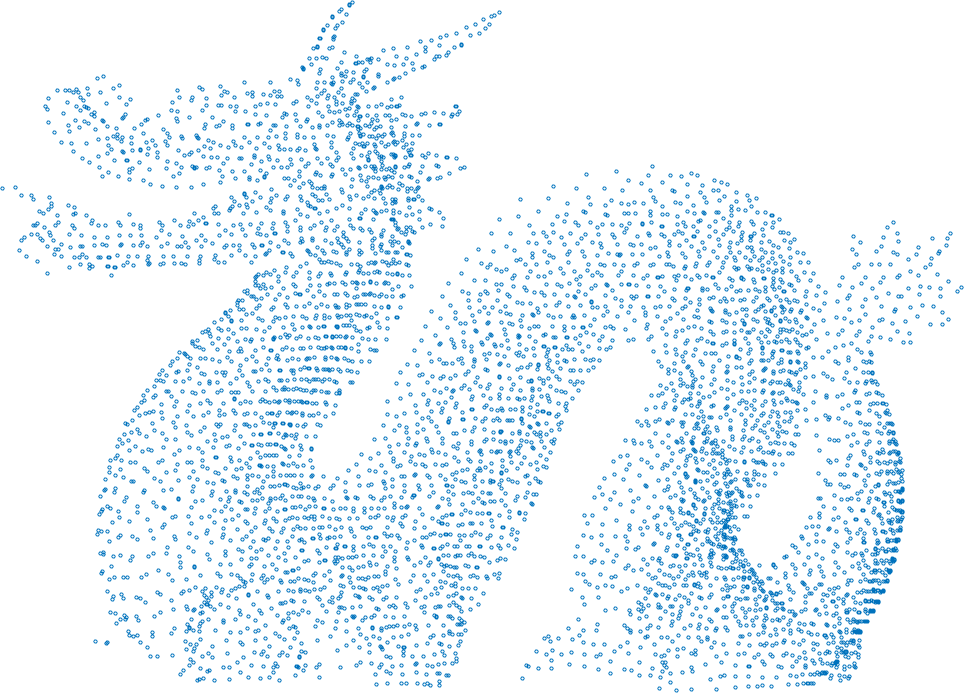}\\
JRMPC-B & JRMPC-I
\end{tabular}
\end{center}
\caption{The GMM means obtained from JRMPC-based algorithms for (top) ``Bunny'' and (bottom) ``Dragon''  Unlike JRMPC-B, JRMPC-I leads to non-uniformly distributed mixture components (biased towards the initial sets) since ``old'' means cannot be freely re-distributed (\emph{best viewed on-screen}).}
\label{fig:means}
\end{figure}

\begin{table*}
\centering
\caption{Performance of multi-view registration methods when points are perturbed by gaussian noise (SNR: $25$dB).}
\label{tab:multiview-reg-results-noise}
\begin{tabular}{lccccccc}
\hline
~ & Raw-data & Initialization & MV-ICP~\cite{Williams2001} & MAICP~\cite{Govindu2014} & MATrICP~\cite{ZhongyuLi2014} & JRMPC-B & JRMPC-I \\
\hline
\hline
Bunny & 3.45 & 2.42 & 2.66 & 2.87 & 2.37 & \bf{1.07} & 1.41\\
Dragon & 7.28 & 7.34 & 7.37 & 3.28 & 1.55 & \bf{0.64} & 0.89 \\
Happy Buddha & 10.77 & 6.88 & 6.86 & 4.13 & 1.92 & \bf{1.18} & 1.69 \\
\hline
\end{tabular}
\end{table*}

Both JRMPC-B and JRMPC-I consider the same number of components ($K\simeq4000$) while the initial centers are randomly selected points from roughly aligned sets. For JRMPC-I, when a new set appears, $K/M$ components are rejected and re-initialized with points from the new set. This is to enforce the displacement of some GMM means towards the new data, as long as model growing is not considered here. Several conditions may apply to this rejection stage. Here, we first reject degenerate clusters ($\sigma^2=\epsilon^2)$ (if any) and we randomly select old components to replace. One iteration of integration step and $30$ refinement iterations with JRMPC-B are allowed owing to the different viewpoints, while we let the algorithm run $50$ cycles to register the two first sets. Note that the current implementations of MV-ICP and MAICP consider the closed-loop known, that is, pairing the last with the first set, while MAICP also considers the scan boundaries known and rejects such points for potential matching. Instead, both versions of our algorithm as well as MATrICP make no use of any prior knowledge about the loop and the overlap. 
\begin{figure*}
\centering
\begin{tabular}{cccccc}
\multicolumn{6}{c}{\includegraphics[width=0.95\textwidth]{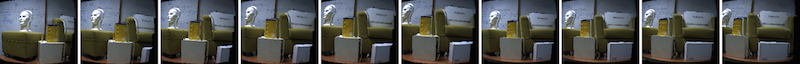}}\\
  \includegraphics[width=0.14\textwidth]{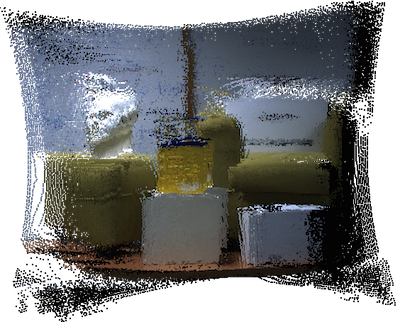} 
&\includegraphics[width=0.14\textwidth]{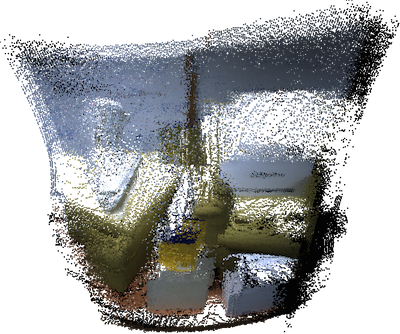}
&\includegraphics[width=0.14\textwidth]{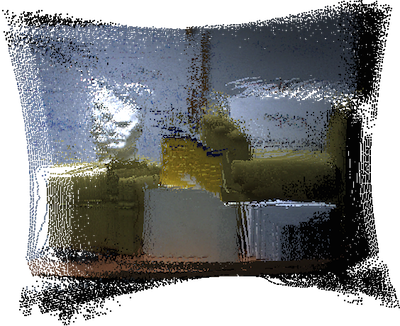}
&\includegraphics[width=0.14\textwidth]{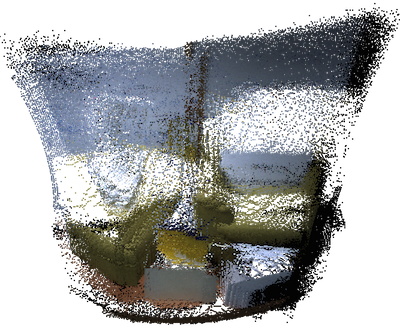}
&\includegraphics[width=0.14\textwidth]{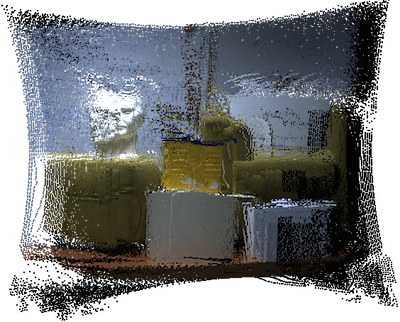}
&\includegraphics[width=0.14\textwidth]{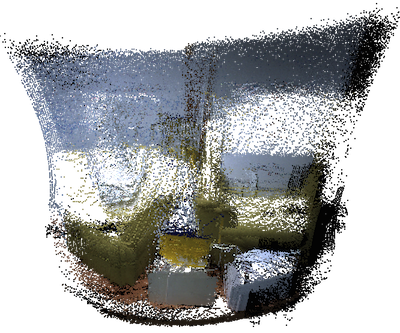}\\
\multicolumn{2}{c}{seqICP} & \multicolumn{2}{c}{MVICP} & \multicolumn{2}{c}{MAICP} \\
\includegraphics[width=0.14\textwidth]{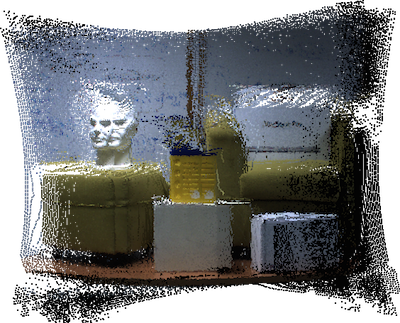}
&\includegraphics[width=0.14\textwidth]{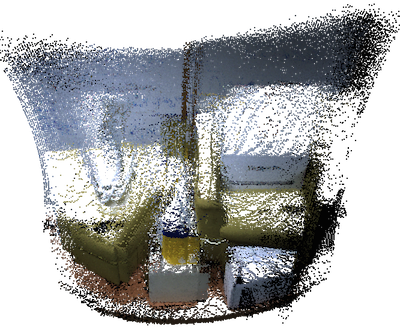}
&\includegraphics[width=0.14\textwidth]{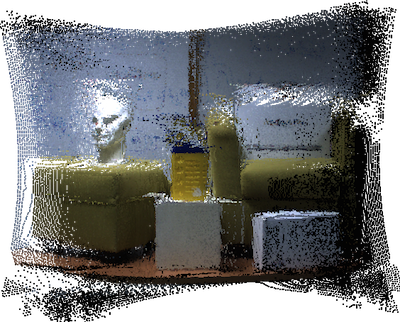}
&\includegraphics[width=0.14\textwidth]{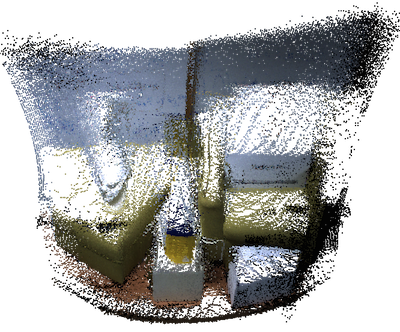}
&\includegraphics[width=0.14\textwidth]{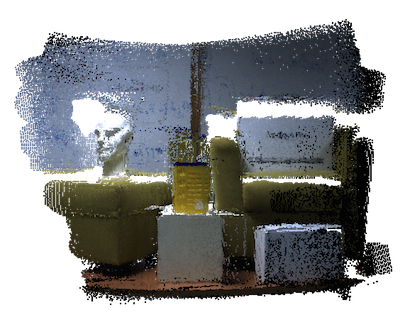}
&\includegraphics[width=0.14\textwidth]{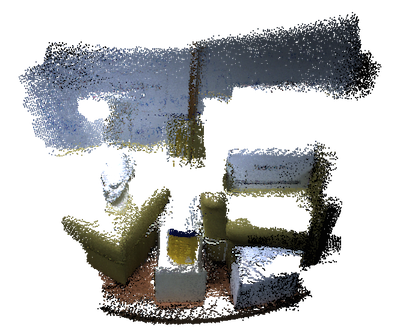}\\
\multicolumn{2}{c}{MATrICP} & \multicolumn{2}{c}{JRMPC-B} & \multicolumn{2}{c}{JRMPC-B after outlier removal} 
\end{tabular}
\caption{Integrated point clouds from the joint registration of $10$ TOF images that record a static scene (EXBI data-set). \emph{Top}: color images that roughly show the scene content of each range image (occlusions due to cameras baseline may cause texture artefacts). \emph{Bottom:} front-view and top-view of integrated sets after joint registration}
\label{fig:real_data}
\end{figure*}

Table~\ref{tab:multiview-reg-results} shows the registration error of the methods.  As expected, MV-ICP fails to provide accurate registration owing to fixed matches. The proposed algorithm along with MATrICP achieve the most accurate registration, while JRMPC-I provides results of sufficient quality. Indeed, as claimed in~\cite{ZhongyuLi2014}, it seems that motion averaging benefits from more robust versions of ICP. The corresponding integrated models of the best performing algorithms are shown in Fig.~\ref{fig:reconstructions}. Likewise, MAICP is less accurate while JRMPC schemes and MATrICP provide very good reconstructions. 	

Fig.~\ref{fig:cross-sections} shows cross-sections of the reconstructions obtained by the proposed and motion-averaging algorithms (best viewed on-screen). The more ``clean'' and solid the sketch, the more accurate the alignment. The algorithms achieve to correct the initial sketch of the pair-wise ICP method. A detailed look verifies the superiority of the proposed batch method and the potential of the incremental version. Note that down-sampling makes short lines intersect in the cross-sections, even when using the ground truth motions. 


Despite its incremental nature, JRMPC-I achieves comparable reconstructions and closes the loop successfully. However, the components are not distributed in the same way. Fig.~\ref{fig:means} shows the distribution of means after running both versions of JRMPC. Despite the refinement step and the rejection stage, the means seem to remain a little biased towards initial sets, which might be problematic with long data sequences. In such a scenario, one should enforce a constraint so that new components that replace the rejected ones entirely belong to new scene surface. Note that detecting the points that may belong to the new part of the scene/object when the depth sensor is moving is easy with today hybrid sensors that deliver visual and inertial data. 

Table~ \ref{tab:multiview-reg-results-noise} provides a quantitative comparison between the methods when the point-sets are further perturbed with noise of  SNR=$25$dB. As seen, the motion averaging methods seem to be more sensitive than the proposed ones. This is mainly because the GMM means get cleaned over time and the registration module in JRMPC is more robust to noise. As a consequence, even JRMPC-I outperforms the motion averaging methods. The presence of noise make the illustration of cross-sections and integrated models meaningless.

Remarkably, we experimentally found that fixing the variance for the initial iterations make JRMPC-B converge at a lower level. When the sets are roughly aligned, a fixed and reasonable value of the variance (that make each cluster include a few points) leads to better distributed means in terms of the object skeleton, which in turn lead to more accurate transformations. This is because the skeleton carries more informative points than the surface itself. Then, the update of the variance leads to better reconstruction of the object and to ``safe'' refinements of the rotations. From a mathematical point of view, this strategy helps avoiding local minima in the variance-rotations subspace.

\subsection{Real Data}
In~\cite{Evangelidis-ECCV-2014}, we tested JRMPC-B along with pairwise strategies on EXBI dataset, that contains depth data captured with a time-of-flight camera rigidly attached to two color cameras. Once calibrated \cite{Hansard2014,Hansard2015}, this TOF-stereo sensor provides RGB-D data. The EXBI data consist of ten point  clouds gathered by manually moving the TOF-stereo sensor in front of a scene, e.g., Fig.~\ref{fig:real_data}. Each point cloud contains approximatively $25,000$ points. While JRMPC-B only uses the depth data, color information is used the final assessment and also shows the potential for fusing RGB-D data. 

The comparison in~\cite{Evangelidis-ECCV-2014} showed that, unlike JRMPC, all the pairwise strategies suffer from misalignments and need further processing, e.g., motion averaging.\footnote{we also refer the reader to the supplementary material of ~\cite{Evangelidis-ECCV-2014}} Therefore, we test the performance of MAICP, MATrICP and MVICP  on EXBI data-set and compare with JRMPC. SeqICP is used to roughly initialize the transformations of the point clouds. 

Fig.~\ref{fig:real_data} shows the front and top view of the integrated sets obtained by seqICP (initialization) as well as by MVICP, MAICP, MATrICP and JRMPC algorithms. Both versions of the proposed algorithm provide visually similar results. As verified, the motion averaging method cannot fully compensate for the misalignments of the initialization. This is shown even in front views, e.g., on the dummy head area. Again, MATrICP is more robust than MAICP, while MVICP clearly underperforms. The proposed scheme, however, achieves to register the point clouds accurately. Despite the large number of outliers, we are also able to get an outlier-free reconstruction of the scene based on the above thresholding principle. Of particular note, finally, is that JRMPC obtains these results with only $450$ components, a fact that further validates its potential.

Finally, we evaluate the performance JRMPC-I with a large number of point clouds collected with a moving sensor, namely the TUM dataset \cite{sturm12iros}. In particular, we converted the depth sequences \emph{fr1$/$desk} and 
\emph{fr2$/$desk} from this dataset into two sequences of $570$ and $2880$ point sets, respectively. The first sequence includes several sweeps (local loops) over four desks in a typical office environment while the second  sequence includes a full loop around a desk. The sequences are captured with a Kinect and ground-truth camera poses are provided with the help of a motion-capture system. As in the previous experiment, color data are not used by the registration algorithm.

To enable the algorithm to deal with a large number of sets, we considered a hierarchical scheme with two modules, a front-end and a back-end module. The front-end registers groups of $N_f$ successive point-sets  and provides an outlier-free GMM, whose means are referred to as the \emph{mean set}. The back-end module uses JRMPC-I on a temporal window of $N_b$ mean sets, that is, a new mean set is integrated at every $N_f$ times tamps and the local model instance of the window is refined with the batch method. We noticed that applying JRMPC-B on a small number of temporally non-overlapping integrated sets, e.g. one downsampled registered set per $100$ point sets, further improves the joint alignment.

All the initial point sets have been downsampled by a factor of $50$ before running the algorithm. We used $N_b = 10$, while $N_f=3$ and $N_f=10$ for \emph{fr1$/$desk} and \emph{fr2$/$desk}, respectively owing to differences in motion patterns. The overlap between successive groups in the front-end module is one set. The number of components is $3000$ and $6000$ for the back-end  and front-end modules, respectively. The batch method refines the window model for $50$ iterations owing to its relatively small window. 
An optimized implementation of this procedure may lead to real-time performance.

\newcolumntype{C}[1]{>{\centering\arraybackslash}p{#1}}
\begin{table*}[t]
\centering
\caption{RMSE ($m$) of translation for SLAM methods and for the proposed method for the TUM dataset.}
\label{tab:kinect_dataset}
\begin{tabular}{lcccc}
\hline
~  & ORB-SLAM2 (RGB-D)~\cite{Mur-Artal2016-ORB-SLAM2}& Elastic fusion~\cite{Whelan2016}  & RGBD SLAM~\cite{Endres2014} & JRMPC-I  \\
\hline
\hline
\emph{fr1/desk}  & 0.016 & 0.020 & 0.026 & 0.047\\
\emph{fr2/desk}  & 0.009 & 0.071 & 0.057 & 0.034\\
\hline
\end{tabular}
\end{table*}
Table~\ref{tab:kinect_dataset} shows the performance of the proposed algorithm based on the protocol of~\cite{sturm12iros}. Typically, the RMSE of the translation is used for evaluating SLAM methods. The error of JRMPC is computed for all frames while the other algorithms use only keyframes. We also provide the error of state-of-the-art RGB-D SLAM methods~\cite{Mur-Artal2016-ORB-SLAM2, Endres2014, Whelan2016} as a reference (a direct comparison is not fair), as reported in~\cite{Mur-Artal2016-ORB-SLAM2}. Although SLAM methods use both modalities (RGB and depth) and invoke several modules to achieve accurate camera localization, the performance of the proposed algorithm is quite close to theirs. 

\begin{figure}[t!]
\centering
\begin{tabular}{cc}
  \includegraphics[width=0.45\columnwidth]{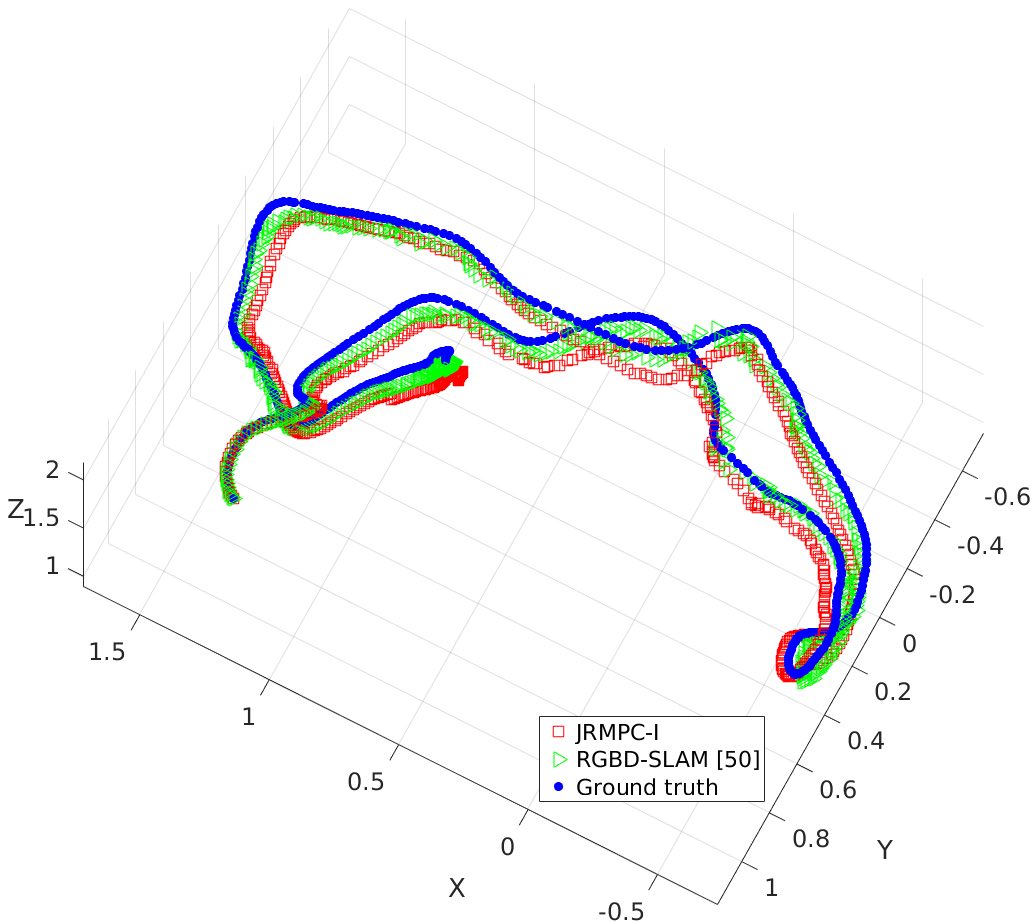} 
&  \includegraphics[width=0.45\columnwidth]{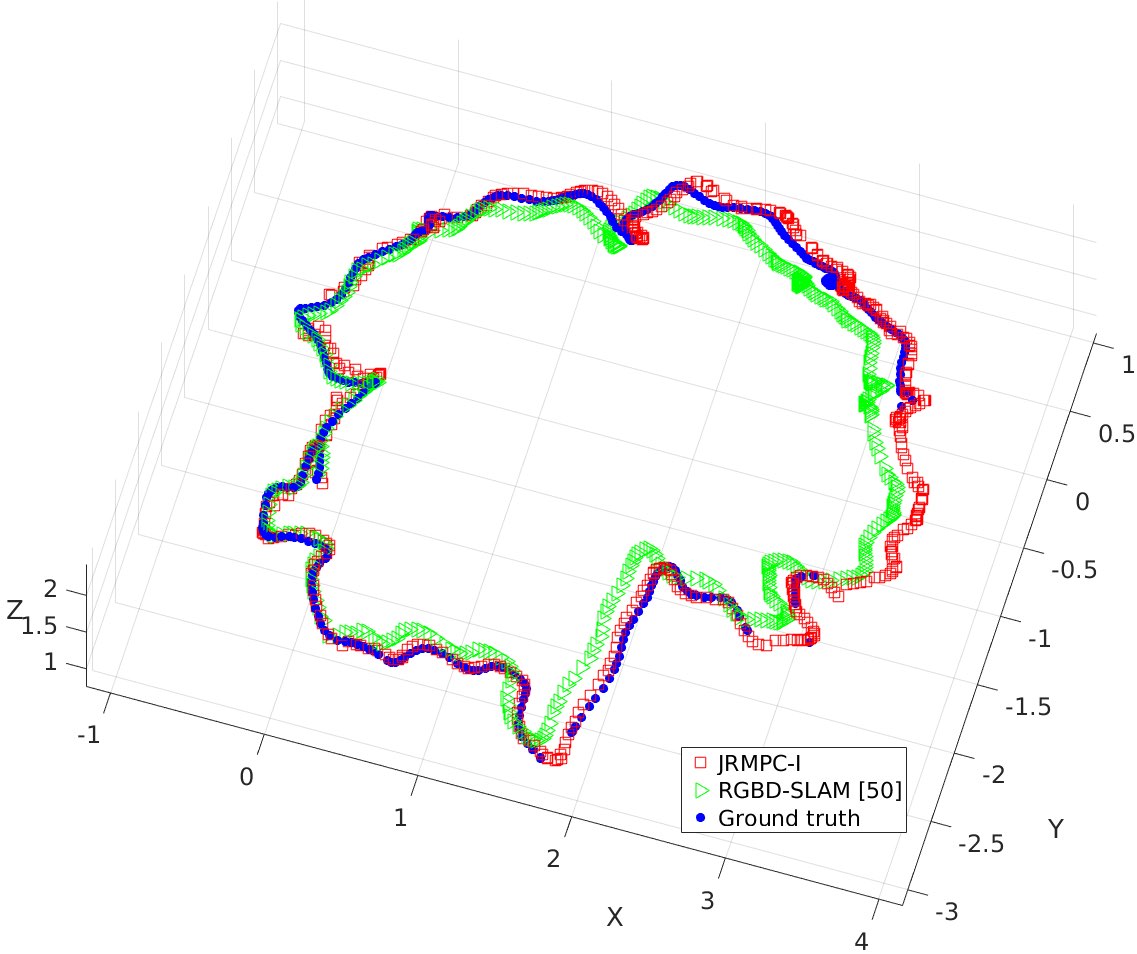}\\
 \emph{fr1/desk} & \emph{fr2/desk} \\
\end{tabular}
\caption{Camera trajectories obtained from JRMPC-I and RGBD-SLAM~\cite{Endres2014}.}
\label{fig:TUM_camera_trajectories}
\end{figure}

\begin{figure}[t!]
\centering
\begin{tabular}{cc}
  \includegraphics[width=0.45\columnwidth]{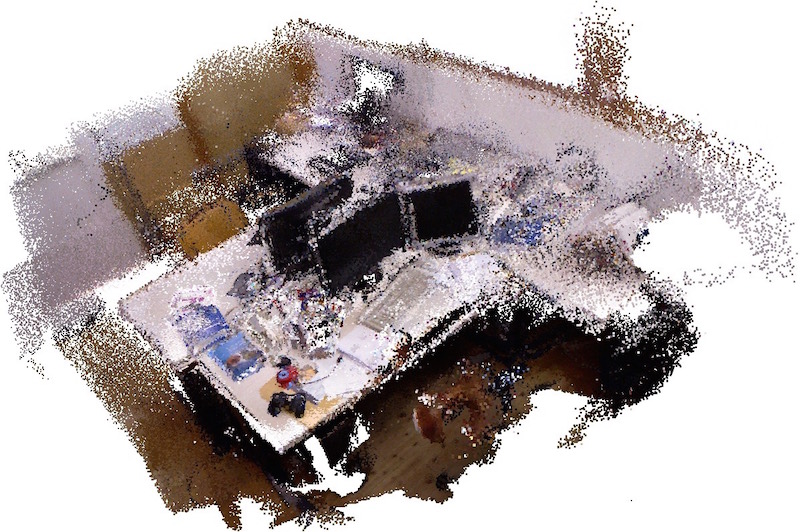} 
&  \includegraphics[width=0.45\columnwidth]{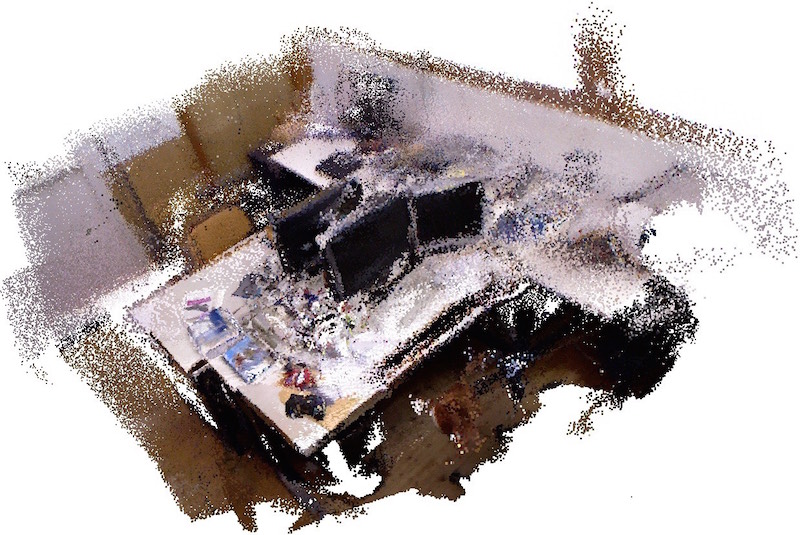}\\
   \includegraphics[width=0.45\columnwidth]{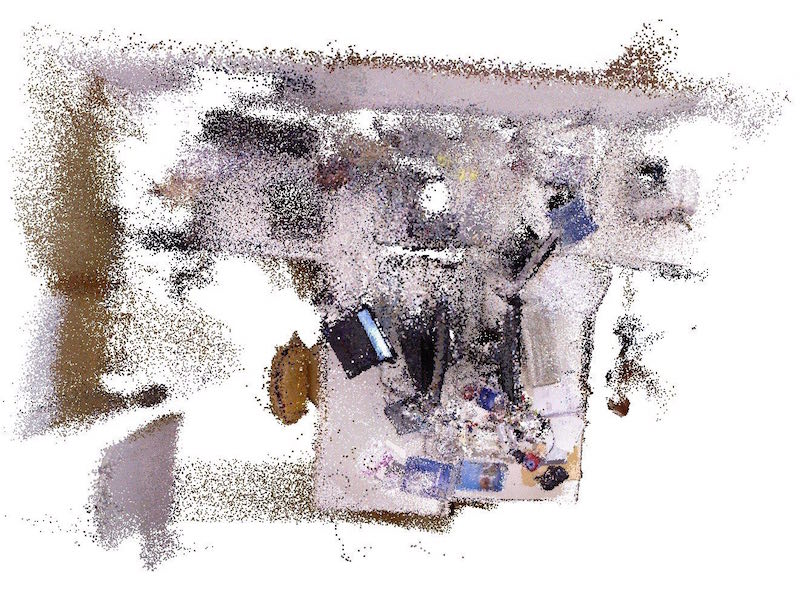} 
&\includegraphics[width=0.45\columnwidth]{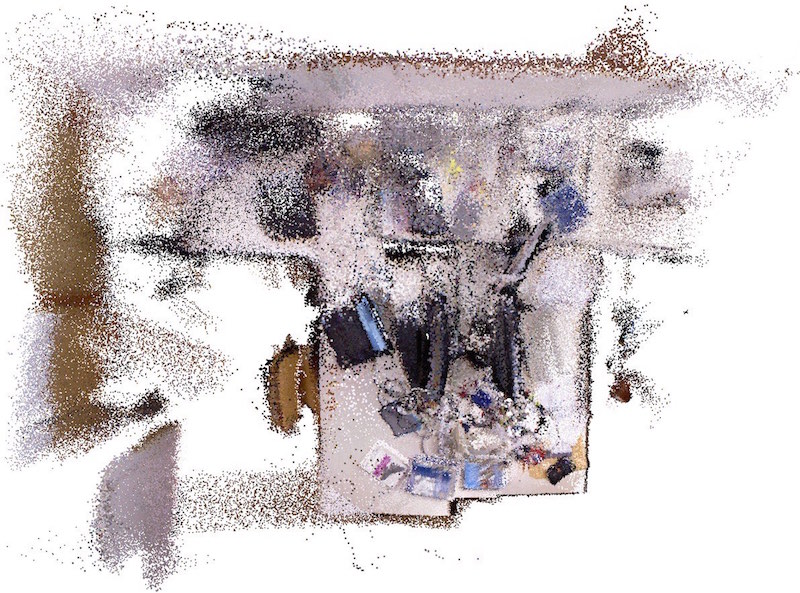}\\
 JRMPC-I & Ground-truth \\
\end{tabular}
\caption{Dense point-cloud reconstruction obtained from JRMPC-I for the sequence \emph{fr1/desk} 
}
\label{fig:TUM_dataset}
\end{figure}

Fig.~\ref{fig:TUM_camera_trajectories} shows the camera trajectories obtained with the proposed algorithm as well as from RGBD-SLAM~\cite{Endres2014}. SLAM methods generally provide smooth trajectories owing to their internal tracking module and pose graph optimization. Instead, our algorithm simply registers depth data in a model-to-frame manner and one may observe local perturbations.  Fig.~\ref{fig:TUM_dataset} shows the final alignment for the \emph{fr1/desk} sequence obtained with JRMPC-I and the corresponding ground truth. Interestingly, the proposed scheme delivers promising reconstructions despite the fact that it uses only depth data without pose graph optimization.

\color{black}

\color{black}


\section{Conclusions}\label{sec:conclusions}
We presented a probabilistic generative model and its associated algorithm to jointly register multiple point sets. The vast majority of state-of-the-art techniques select one of the sets as the model and attempt to align the other sets onto this model. Instead, the proposed method treats all the point sets on an equal footing: any point is considered as realization of a single GMM and the registration is cast into a clustering problem. We formally derived an expectation-maximization algorithm that estimates the GMM parameters as well as the rotations and translations between each individual set and the initially unknown GMM means. An incremental version of the algorithm that efficiently integrates new point sets into the registration pipeline was also derived. We thoroughly validated the proposed method on challenging data sets gathered with depth cameras, we compared it with several state-of-the-art methods, and we showed its potential for effectively fusing depth data. In the future we plan to investigate the use of more efficient representations of generative models, e.g.,~\cite{EckartCVPR2016} and an incremental registration method allowing the number of clusters to grow.

\bibliographystyle{IEEEtran}

\end{document}